\documentclass[twocolumn,10pt]{article}

\usepackage[letterpaper,margin=1in]{geometry}
\usepackage{url}
\usepackage{color}
\usepackage[dvipsnames]{xcolor}
\usepackage{amsmath, amssymb, epsfig}
\usepackage{mathrsfs}
\usepackage{subcaption}
\usepackage{bbm}
\usepackage{color}
\usepackage{appendix}
\usepackage{setspace}
\usepackage{comment}
\usepackage{nccmath}
\DeclareMathOperator*{\argmin}{argmin}
\DeclareMathOperator*{\argmax}{argmax}
\usepackage{ntheorem}
\usepackage{tikz}

\newcommand{\R}{\ensuremath{\mathbb{R}}}

\usepackage{algorithm}
\usepackage[noend]{algorithmic}
\usepackage{comment}
\usepackage{titling}
\usepackage{multirow}
\usepackage{tabularx}

\newcommand{\pars}[1]{\left(#1\right)}
\newcommand{\bracs}[1]{\left[#1\right]}
\newcommand{\mat}[1]{\mathbf{#1}}

\newcommand*\samethanks[1][\value{footnote}]{\footnotemark[#1]}
\thanksmarkseries{arabic}

\newtheorem*{mle}{Maximum Likelihood Estimators}

\providecommand{\keywords}[1]{\noindent\textbf{\small Keywords:} #1}

\begin{document}

\title{\Large Semi-supervised NMF Models for Topic Modeling in Learning Tasks\thanks{JH and DN were partially supported by NSF DMS $\#2011140$ and NSF BIGDATA $\#1740325$. ES was supported by the Moore-Sloan Foundation. Funding from ICERM and the NSF-AWM ADVANCE grant initiated the collaboration.}}
\author{Jamie Haddock\thanks{UCLA, \{\texttt{jhaddock}, \texttt{tmerkh}, \texttt{deanna}\}\texttt{@math.ucla.edu}.}
\and Lara Kassab\thanks{Colorado State University, \texttt{lara.kassab@colostate.edu}.}
\and Sixian Li\thanks{UIUC, \texttt{sixianl2@illinois.edu}.}
\and Alona Kryshchenko\thanks{CSU Channel Islands, \texttt{alona.kryshchenko@csuci.edu}.}
\and Rachel Grotheer\thanks{Wofford College, \texttt{grotheerre@wofford.edu}.}
\and Elena Sizikova\thanks{NYU, \texttt{es5223@nyu.edu}.}
\and Chuntian Wang\thanks{University of Alabama,  \texttt{cwang27@ua.edu}.}
\and Thomas Merkh\samethanks[2]
\and R.\ W.\ M.\ A.\ Madushani\thanks{Boston Medical Center, \texttt{madushani.rajapaksha@bmc.org}.}
\and Miju Ahn\thanks{Southern Methodist University, \texttt{mijua@smu.edu}.}
\and Deanna Needell\samethanks[2] 
\and Kathryn Leonard\thanks{Occidental College, \texttt{leonardk@oxy.edu}.}
}

\date{}

\maketitle

\begin{abstract} {\small We propose several new models for semi-supervised nonnegative matrix factorization (SSNMF) and provide motivation for SSNMF models as maximum likelihood estimators given specific distributions of uncertainty.  We present multiplicative updates training methods for each new model, and demonstrate the application of these models to classification, although they are flexible to other supervised learning tasks.
We illustrate the promise of these models and training methods on both synthetic and real data, and achieve high classification accuracy on the 20 Newsgroups dataset.}\end{abstract}

\vspace{0.2cm}

\section{Introduction}
Frequently, one is faced with the problem of performing a (semi-)supervised learning task on extremely high-dimensional data which contains redundant information.  A common approach is to first apply a dimensionality-reduction or feature extraction technique (e.g., PCA \cite{pearson1901liii}), and then train the model for the learning task on the new, learned 
representation of the data.  One problematic aspect of this two-step approach is that the learned representation of the data may provide ``good" fit, but could suppress data features which are integral to the learning task \cite{guyon2003introduction}.
For this reason, supervision-aware dimensionality-reduction models have become increasingly important in data analysis; such models aim to use supervision in the process of learning the lower-dimensional representation, or even learn this representation alongside the supervised learning model \cite{wang2014role,blei2003latent,pritchard2000inference}. 

In this work, we propose new semi-supervised nonnegative matrix factorization (SSNMF) formulations which provide a dimensionality-reducing topic model and a model for a supervised learning task.   
Our contributions are:
\begin{itemize}
    \item we motivate these proposed SSNMF models and that of~\cite{lee2009semi} as maximum likelihood estimators (MLE) given specific models of uncertainty in the observations;
    \item we derive multiplicative updates for the proposed models that allow for missing data and partial supervision;
    \item we perform experiments on synthetic and real data which illustrate the promise of these models in both topic modeling and supervised learning tasks; and
    \item we demonstrate the promise of SSNMF models for classification relative to the performance of other classifiers on a common benchmark data set.
\end{itemize}

\subsection{Organization}\label{sec:organization}
The paper is organized as follows.  We begin by briefly defining 
notation in Section~\ref{sec:notation} and provide a preliminary review of the fundamental models, nonnegative matrix factorization (NMF)~\cite{lee1999learning} and SSNMF~\cite{lee2009semi} in Section~\ref{sec:preliminaries}, other related work in Section~\ref{sec:related work}, and 
the proposed models in Section~\ref{sec:overview}.  
We motivate the proposed models and that of~\cite{lee2009semi} via MLE in Section~\ref{subsec:MLE}, present the multiplicative update methods for training in Section~\ref{subsec:propmodels}, and present details of a framework for classification with these models in Section~\ref{sec:classification}. We present experimental evidence illustrating the promise of the SSNMF models in Section~\ref{sec: numerical experiments},  
including experiments on synthetic data in Section~\ref{subsec:syntheticdata} which are motivated by the MLE of Section~\ref{subsec:MLE}, and experiments on the 20 Newsgroup dataset in Section~\ref{subsec:20newsdata}.  Finally, we end with some conclusions and discussion of future work in Section~\ref{sec:conclusion}.

\subsection{Notation}\label{sec:notation}
Our models make use of two matrix similarity measures.  The first is the standard Frobenius norm, $\|\mat{A} - \mat{B}\|_F$. 
The second is the \emph{information divergence} or I-divergence, a measure defined between nonnegative matrices $\mat A$ and $\mat B$,
\begin{equation}\label{def}
    D(\mat{A} \| \mat{B}) = \sum_{i,j} \bigg( \mat{A}_{ij} \log{\frac{\mat{A}_{ij}}{\mat{B}_{ij}}} - \mat{A}_{ij} + \mat{B}_{ij}\bigg),
\end{equation}
where $D(\mat{A} \| \mat{B}) \geq 0$ with equality if and only if $\mat{A} = \mat{B}$~\cite{lee2001algorithms}.
Because the information divergence reduces to the Kullback-Leibler divergence when $\mat{A}$ and $\mat{B}$ represent probability distributions, i.e., $\sum \mat{A}_{ij} = \sum \mat{B}_{ij} = 1$, it is often referred to as the generalized Kullback-Leibler divergence~\cite{favaro20073}.
 
In the following, $\mat{A}/\mat{B}$ indicates element-wise division, $\mat A \odot \mat B$ indicates element-wise multiplication, and $\mat A \mat B$ denotes standard matrix multiplication.  We denote the set of non-zero indices of a matrix by $\text{supp}(\mat A) := \{(i,j) : \mat A_{ij}\ne 0\}$.  When an $n_1\times n_2$ matrix is to be restricted to have only nonnegative entries, we write $\mat A\geq 0$ and $\mat A\in \mathbb{R}^{n_1 \times n_2}_{\ge 0}$. We let $\mathbf{1_k}$ denote the length-$k$ vector consisting of ones, $ \mathbf{1_k} =  \begin{bmatrix}  1 , \cdots 1 \end{bmatrix}^\top \in \R^k$, and similarly $\mathbf{0_k}$ denotes the vector of all zeros, $\mathbf{0_k} =  \begin{bmatrix}  0 , \cdots 0 \end{bmatrix}^\top \in \R^k$.

We let $\mathcal{N}\pars{z \middle| \mu,\sigma^2}$ denote the Gaussian density function for a 
random variable $z$ with mean $\mu$ and variance $\sigma^2$, 
and $\mathcal{PO}\pars{z \middle| \nu}$ denotes the Poisson density function for 
a random variable $z$ with nonnegative intensity parameter $\nu$.

\subsection{Preliminaries}
\label{sec:preliminaries}
In this section, we give a brief overview of the NMF and SSNMF methods.

\subsubsection*{Nonnegative Matrix Factorization}
Given a nonnegative matrix $\mat X \in \mathbb{R}^{n_1 \times n_2}_{\ge 0}$ and a target dimension $r \in \mathbb N$, NMF decomposes $\mat X$ into a product of two low-dimensional nonnegative matrices.
The model seeks $\mat A$ and $\mat S$ so that $\mat X \approx \mat A \mat S$,
where $\mat A \in \mathbb{R}^{n_1 \times r}_{\ge 0}$ is called the dictionary matrix and $\mat S\in \mathbb{R}^{r \times n_2}_{\ge 0}$ is called the representation matrix.
Typically, $r$ is chosen such that $r<\min\{n_1,n_2\}$ to reduce the dimension of the original data matrix or reveal latent themes in the data.
Data points are typically stored as columns of $\mat X$, thus $n_1$ represents the number of features, and $n_2$ represents the number of samples.
The columns of $\mat A$ are generally referred to as \emph{topics}, which are characterized by features of the data set.
Each column of $\mat S$ provides the approximate representation of the respective column in $\mat X$ in the lower-dimensional space spanned by the columns of $\mat A$.
Thus, the data points are well approximated by an additive linear combination of the latent topics.

Several formulations for this nonnegative approximation, $\mat X \approx \mat A \mat S$, have been studied~\cite{cichocki2009nonnegative,lee1999learning,lee2001algorithms,yang2011kullback}; 
e.g.,
\begin{equation}\label{eq:energy frobenius}
\argmin \limits_{\mat A\geq 0,\mat S\geq 0} \|\mat X - \mat A \mat S\|_F^2 \;\text{ and }
\argmin \limits_{\mat A\geq 0, \mat S\geq 0} D(\mat X\| \mat A \mat S),
\end{equation}
where $D(\cdot\|\cdot)$ is the information divergence defined in~\eqref{def}.
In what follows, we refer to the left formulation of \eqref{eq:energy frobenius} as $\|\cdot\|_F$-NMF and the right formulation of \eqref{eq:energy frobenius} as $D(\cdot\|\cdot)$-NMF. We refer the reader to~\cite{cichocki2009nonnegative} for discussions of similarity measures and generalized divergences (where information divergence is a particular case), and~\cite{li2012fast,sra2006generalized} for generalized nonnegative matrix approximations with Bregman divergences.

Multiplicative update algorithms for both formulations of~\eqref{eq:energy frobenius} have been proposed~\cite{lee1999learning, lee2001algorithms}.
These algorithms are widely adopted because they are easy to implement, do not require
user-specified hyperparameters, preserve the nonnegativity constraints, and have desirable monotonicity properties \cite{lee2001algorithms}. Other popular algorithms include projected gradient descent and alternating least-squares \cite{cichocki2009nonnegative,kim2008fast,kim2008nonnegative,lin2007projected}.

NMF has gained popularity recently due to large scale data demands of
applications 
such as document clustering \cite{gaussier2005relation,shahnaz2006document,xu2003document,berry2005email,pauca2004text}, 
image processing~\cite{guillamet2002non,hoyer2002non,lee1999learning}, 
financial data mining~\cite{de2008analysis},  
audio processing \cite{cichocki2006new,gemmeke2013exemplar}, 
and genetics \cite{liu2017regularized}. 

\subsubsection*{Semi-supervised NMF}
\begin{figure}[t]
    \centering
    \definecolor{zzttqq}{rgb}{0.6,0.2,0.}
\definecolor{cczzqq}{rgb}{0.8,0.6,0.}
\definecolor{qqttqq}{rgb}{0.,0.2,0.}
\definecolor{qqttzz}{rgb}{0.,0.2,0.6}
\definecolor{yqqqqq}{rgb}{0.5019607843137255,0.,0.}
\definecolor{ubqqys}{rgb}{0.29411764705882354,0.,0.5098039215686274}
\begin{tikzpicture}[line cap=round,line join=round,x=1.0cm,y=1.0cm,scale=0.55]
\clip(-4.0550855952123235,-1.559258692596843) rectangle (11.063424250554109,5.0585794749534295);
\fill[line width=1.pt,color=ubqqys,fill=ubqqys,fill opacity=0.10000000149011612] (-4.,5.) -- (1.,5.) -- (1.,2.) -- (-4.,2.) -- cycle;
\fill[line width=1.pt,color=yqqqqq,fill=yqqqqq,fill opacity=0.10000000149011612] (3.,5.) -- (3.,2.) -- (5.,2.) -- (5.,5.) -- cycle;
\fill[line width=1.pt,color=qqttzz,fill=qqttzz,fill opacity=0.10000000149011612] (6.,5.) -- (6.,3.) -- (11.,3.) -- (11.,5.) -- cycle;
\fill[line width=1.pt,color=qqttqq,fill=qqttqq,fill opacity=0.10000000149011612] (-4.,1.) -- (-4.,-0.4) -- (1.,-0.4) -- (1.,1.) -- cycle;
\fill[line width=1.pt,color=cczzqq,fill=cczzqq,fill opacity=0.10000000149011612] (3.,1.) -- (3.,-0.4) -- (5.,-0.4) -- (5.,1.) -- cycle;
\fill[line width=1.pt,color=qqttzz,fill=qqttzz,fill opacity=0.10000000149011612] (6.,1.) -- (6.,-1.) -- (11.,-1.) -- (11.,1.) -- cycle;
\fill[line width=1.pt,color=zzttqq,fill=zzttqq,fill opacity=0.10000000149011612] (10.00269048257823,5.) -- (10.4,5.) -- (10.40109499739337,3.) -- (10.00269048257823,3.) -- cycle;
\fill[line width=1.pt,color=zzttqq,fill=zzttqq,fill opacity=0.10000000149011612] (10.00269048257823,1.) -- (10.4,1.) -- (10.4,-1.) -- (10.,-1.) -- cycle;
\fill[line width=1.pt,color=zzttqq,fill=zzttqq,fill opacity=0.10000000149011612] (0.,5.) -- (0.4,5.) -- (0.4,2.) -- (0.,2.) -- cycle;
\fill[line width=1.pt,color=zzttqq,fill=zzttqq,fill opacity=0.10000000149011612] (0.,1.) -- (0.4,1.) -- (0.4,-0.4) -- (0.,-0.4) -- cycle;
\draw [line width=1.pt,color=ubqqys] (-4.,5.)-- (1.,5.);
\draw [line width=1.pt,color=ubqqys] (1.,5.)-- (1.,2.);
\draw [line width=1.pt,color=ubqqys] (1.,2.)-- (-4.,2.);
\draw [line width=1.pt,color=ubqqys] (-4.,2.)-- (-4.,5.);
\draw [line width=1.pt,color=yqqqqq] (3.,5.)-- (3.,2.);
\draw [line width=1.pt,color=yqqqqq] (3.,2.)-- (5.,2.);
\draw [line width=1.pt,color=yqqqqq] (5.,2.)-- (5.,5.);
\draw [line width=1.pt,color=yqqqqq] (5.,5.)-- (3.,5.);
\draw [line width=1.pt,color=qqttzz] (6.,5.)-- (6.,3.);
\draw [line width=1.pt,color=qqttzz] (6.,3.)-- (11.,3.);
\draw [line width=1.pt,color=qqttzz] (11.,3.)-- (11.,5.);
\draw [line width=1.pt,color=qqttzz] (11.,5.)-- (6.,5.);
\draw [line width=1.pt,color=qqttqq] (-4.,1.)-- (-4.,-0.4);
\draw [line width=1.pt,color=qqttqq] (-4.,-0.4)-- (1.,-0.4);
\draw [line width=1.pt,color=qqttqq] (1.,-0.4)-- (1.,1.);
\draw [line width=1.pt,color=qqttqq] (1.,1.)-- (-4.,1.);
\draw [line width=1.pt,color=cczzqq] (3.,1.)-- (3.,-0.4);
\draw [line width=1.pt,color=cczzqq] (3.,-0.4)-- (5.,-0.4);
\draw [line width=1.pt,color=cczzqq] (5.,-0.4)-- (5.,1.);
\draw [line width=1.pt,color=cczzqq] (5.,1.)-- (3.,1.);
\draw [line width=1.pt,color=qqttzz] (6.,1.)-- (6.,-1.);
\draw [line width=1.pt,color=qqttzz] (6.,-1.)-- (11.,-1.);
\draw [line width=1.pt,color=qqttzz] (11.,-1.)-- (11.,1.);
\draw [line width=1.pt,color=qqttzz] (11.,1.)-- (6.,1.);
\draw (-2.0025738424860255,3.977294438567234) node[anchor=north west] {$\mat X$};
\draw (3.505396044010993,3.977294438567234) node[anchor=north west] {$\mat A$};
\draw (8.098152386062583,4.499134513077385) node[anchor=north west] {$\mat S$};
\draw (-2.0025738424860255,0.7197516409610733) node[anchor=north west] {$\mat Y$};
\draw (3.505396044010993,0.7197516409610733) node[anchor=north west] {$\mat B$};
\draw (8.098152386062583,0.4885249246209731) node[anchor=north west] {$\mat S$};
\draw (1.5033734699352812,3.927294438567234) node[anchor=north west] {$\approx$};
\draw (1.5033734699352812,0.6697516409610733) node[anchor=north west] {$\approx$};
\draw (-2.8478494766911844,2.1727563420515913) node[anchor=north west] {$n_1 \times n_2$};
\draw (2.903249290415817,2.1727563420515913) node[anchor=north west] {$n_1 \times r$};
\draw (7.593490110027475,3.1570498492419424) node[anchor=north west] {$r \times n_2$};
\draw (-2.554107237911151,-0.3011063065342692) node[anchor=north west] {$k \times n_2$};
\draw (3.000427088890859,-0.3011063065342692) node[anchor=north west] {$k \times r$};
\draw (7.593490110027475,-0.8269952989094781) node[anchor=north west] {$r \times n_2$};
\draw [line width=1.pt,color=zzttqq] (10.00269048257823,5.)-- (10.4,5.);
\draw [line width=1.pt,color=zzttqq] (10.4,5.)-- (10.40109499739337,3.);
\draw [line width=1.pt,color=zzttqq] (10.40109499739337,3.)-- (10.00269048257823,3.);
\draw [line width=1.pt,color=zzttqq] (10.00269048257823,3.)-- (10.00269048257823,5.);
\draw [line width=1.pt,color=zzttqq] (10.00269048257823,1.)-- (10.4,1.);
\draw [line width=1.pt,color=zzttqq] (10.4,1.)-- (10.4,-1.);
\draw [line width=1.pt,color=zzttqq] (10.4,-1.)-- (10.,-1.);
\draw [line width=1.pt,color=zzttqq] (10.,-1.)-- (10.00269048257823,1.);
\draw [line width=1.pt,color=zzttqq] (0.,5.)-- (0.4,5.);
\draw [line width=1.pt,color=zzttqq] (0.4,5.)-- (0.4,2.);
\draw [line width=1.pt,color=zzttqq] (0.4,2.)-- (0.,2.);
\draw [line width=1.pt,color=zzttqq] (0.,2.)-- (0.,5.);
\draw [line width=1.pt,color=zzttqq] (0.,1.)-- (0.4,1.);
\draw [line width=1.pt,color=zzttqq] (0.4,1.)-- (0.4,-0.4);
\draw [line width=1.pt,color=zzttqq] (0.4,-0.4)-- (0.,-0.4);
\draw [line width=1.pt,color=zzttqq] (0.,-0.4)-- (0.,1.);
\end{tikzpicture}
    \caption{Given the number of classes $k$, and a desired dimension $r$,
    SSNMF is formulated as a joint factorization of a data matrix $\mat X \in \R_{\ge 0} ^{n_1 \times n_2}$ and a label matrix $\mat Y \in \R_{\ge 0} ^{k \times n_2}$, sharing representation factor $\mat S \in \R_{\ge 0} ^{r \times n_2}$.} 
    \label{fig:SSNMF_illustration}
\end{figure}
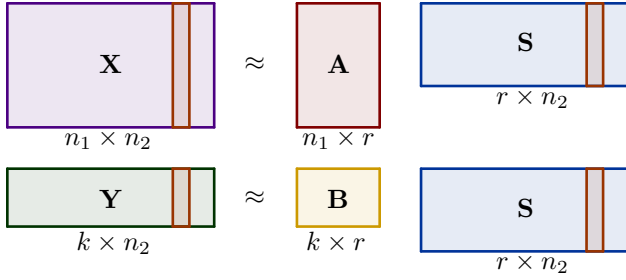

SSNMF is a modification of NMF to jointly incorporate a data matrix and a (partial) class label matrix. 
Given a data matrix $\mat X \in \R_{\ge 0}^{n_1 \times n_2}$ and a class label matrix $\mat Y \in \R_{\ge 0}^{k \times n_2}$, SSNMF is defined by 
\begin{equation} \label{eqn:ssnmf}
   \argmin\limits_{\mat A, \mat S, \mat B\geq 0} \underbrace{\|\mat W \odot(\mat X - \mat A \mat S)\|_F^2}_\text{Reconstruction Error} + \lambda \underbrace{\|\mat L \odot(\mat Y - \mat B \mat S)\|_F^2}_\text{Classification Error},
\end{equation}
where $\quad \mat A \in \R^{n_1 \times r}_{\ge 0},\, \mat B \in \R^{k \times r}_{\ge 0},\, \mat S \in \R^{r \times n_2}_{\ge 0}$, and the regularization parameter $\lambda >0$ governs the relative importance of the supervision term~\cite{lee2009semi}. See Figure~\ref{fig:SSNMF_illustration} for an illustration of the SSNMF model.
We denote this objective function as $F_1(\mat A, \mat B, \mat S; \mat X, \mat Y)$. 
The binary weight matrix $\mat W$ accommodates  missing data by indicating observed and unobserved data entries (that is, $\mat W_{ij} = 1$ if $\mat X_{ij}$ is observed, and $\mat W_{ij} = 0$ otherwise).
 Similarly, $\mat L \in \R^{k \times n_2}$ is a weight matrix that indicates the presence or absence of a label (that is, $\mat L_{:,j} = \mathbf{1_k}$ if the label of $\mat X_{:,j}$ is known, and $\mat L_{:,j} = \mathbf{0_k}$ otherwise).  
Multiplicative updates have been previously developed for SSNMF for the Frobenius norm, and the resulting performance of clustering and classification is improved by incorporating data labels into NMF~\cite{lee2009semi}.
To differentiate this model from the proposed SSNMF models, we refer to the model defined by~\eqref{eqn:ssnmf} as $(\|\cdot\|_F,\|\cdot\|_F)$-SSNMF.

\subsection{Related Work}\label{sec:related work}
In this section, we describe related work most relevant to our own.  This is not meant to be a comprehensive study of these areas.  We focus on work in three main areas: statistical motivation for NMF models, models for simultaneous dimension reduction and supervised learning, and semi-supervised and joint NMF models.

\subsubsection*{Statistical Motivation for NMF} 
The most common discrepancy measures for NMF are the Frobenius norm and the information divergence.  One reason for this popularity is that $\|\cdot\|_F$-NMF and $D(\cdot\|\cdot)$-NMF correspond to the MLE given an assumed latent generative model and a Gaussian and Poisson model of uncertainty, respectively~\cite{cemgil2008bayesian,favaro20073,virtanen2008bayesian}. In \cite{cemgil2008bayesian,virtanen2008bayesian}, the authors go further towards a Bayesian approach, introduce application-appropriate prior distributions on the latent factors, and apply \emph{maximum a posteriori} (MAP) estimation. 
Additionally, 
under certain conditions, $D(\cdot\|\cdot)$-NMF is equivalent to probabilistic latent semantic indexing~\cite{ding2008equivalence}.

\subsubsection*{Dimension Reduction and Learning} 
There has been much work developing dimensionality-reduction models that are supervision-aware. 
Semi-supervised clustering makes use of known label information or other supervision \emph{and} the data features while forming clusters~\cite{basu2002semi,klein2002instance,wagstaff2001constrained}. These techniques generally make use of label information in the cluster initialization or during cluster updating via must-link and cannot-link constraints; empirically, these approaches are seen to increase mutual information between computed clusters and user-assigned labels~\cite{basu2002semi}.
Semi-supervised feature extraction makes use of supervision information in the feature extraction process~\cite{fukumizu2004dimensionality, sheikhpour2017survey}.  These approaches are generally \emph{filter}- or \emph{wrapper}-based approaches, and distinguished by their underlying supervision type~\cite{sheikhpour2017survey}.

\subsubsection*{Semi-supervised and Joint NMF}
Since the seminal work of Lee et al.~\cite{lee2009semi}, semi-supervised NMF models have been studied in a variety of settings. The works~\cite{chen2008non, fei2008semi,jia2019semi} propose models which exploit cannot-link or must-link supervision.  In~\cite{cho2011nonnegative}, the authors introduce a model with information divergence penalties on the reconstruction and on supervision terms which influence the learned factorization to approximately reconstruct coefficients learned before factorization by a support-vector machine (SVM).
Several works~\cite{jia2004fisher,xue2006modified,zafeiriou2006exploiting} propose a supervised NMF model that incorporates Fisher discriminant constraints into NMF for classification.  
Furthermore, joint factorization of two data matrices, like that of SSNMF, is described more generally and denoted Simultaneous NMF in~\cite{cichocki2009nonnegative}.

\subsection{Overview of Proposed Models}\label{sec:overview}
Our proposed models generalize NMF to supervised learning tasks
and provide a topic model which simultaneously provides a lower dimensional representation of the data and a predictive model for targets. 
We denote the data matrix as $\mat X \in \mathbb{R}^{n_1 \times n_2}_{\ge 0}$ and the supervision matrix as $\mat Y \in \mathbb{R}^{k \times n_2}_{\ge 0}$. Following along with~\cite{lee2009semi}, the data observations are the columns of $\mat X$ and the associated targets (e.g., labels) are the columns of $\mat Y$.  Our models seek $\mat A \in \mathbb{R}^{n_1 \times r}_{\ge 0}$, $\mat S \in \mathbb{R}^{r \times n_2}_{\ge 0}$, and $\mat B \in \mathbb{R}^{k \times r}_{\ge 0}$ which jointly factorize $\mat X$ and $\mat Y$; that is $\mat X \approx \mat A \mat S$ and $\mat Y \approx \mat B \mat S$.  We point out the simple fact that these joint factorizations can be stacked into a single NMF (visualized in Figure~\ref{fig:SSNMF_illustration})
\begin{equation}
    \begin{bmatrix} \mat X \\ \mat Y \end{bmatrix} \approx \begin{bmatrix} \mat A \\ \mat B \end{bmatrix} \mat S. \label{eq:stackedNMF}
\end{equation}
In each model, the matrix $\mat A \in \mathbb{R}_{\ge 0}^{n_1 \times r}$ provides a basis for the lower-dimensional space, $\mat S \in \mathbb{R}_{\ge 0}^{r \times n_2}$ provides the coefficients representing the projected data in this space, and $\mat B \in \mathbb{R}_{\ge 0}^{k \times r}$ provides the supervision model which predicts the targets given the representation of points in the lower-dimensional space.  We allow for missing data and labels or confidence-weighted errors via the data-weighting matrix $\mat W \in \mathbb{R}^{n_1 \times n_2}_{\ge 0}$ and the label-weighting matrix $\mat L \in \mathbb{R}^{k \times n_2}_{\ge 0}$. Each resulting joint-factorization model is defined by the error functions applied to the reconstruction and supervision factorization terms. We denote the model 

\begin{equation}\label{eq:RandSSSNMF}
\argmin\limits_{\mat A, \mat S, \mat B\geq 0}\;\, \underbrace{R(\mat W\odot \mat X, \mat W\odot \mat A \mat S)}_\text{\tiny Reconstruction Error} + \; \lambda \underbrace{S(\mat L\odot \mat Y, \mat L \odot \mat B \mat S)}_\text{\tiny Supervision Error}
\end{equation} 
as $(R(\cdot,\cdot),S(\cdot, \cdot))$-SSNMF for specific choices of $R$ and $S$. Here, $R(\cdot,\cdot)$ and $S(\cdot,\cdot)$ are the error functions applied to the reconstruction term and supervision term, respectively. 
For clarity, we include Table~\ref{table:models} below which summarizes existing and proposed models, where each proposed model is of the form~\eqref{eq:RandSSSNMF} for specific choices of error functions $R$ and $S$.
\begin{table}[tb] 
\caption{Overview of NMF and SSNMF models.}\label{table:models}
{ \renewcommand{\arraystretch}{1.15}
\resizebox{\columnwidth}{!}{%
\begin{tabular}{ c c }
            \hline
            Model & Objective \\
            \hline
            $\|\cdot\|_F$-NMF~\cite{lee1999learning} & $\argmin \limits_{\mat A, \mat S\geq 0} \|\mat X - \mat A \mat S\|_F^2$\\
            $D(\cdot\|\cdot)$-NMF~\cite{lee2001algorithms} & $\argmin \limits_{\mat A, \mat S\geq 0} D(\mat X\| \mat A \mat S)$\\
            $(\|\cdot\|_F,\|\cdot\|_F)$-SSNMF~\cite{lee2009semi} & $\argmin \limits_{\mat A, \mat B, \mat S\geq 0} \|\mat W \odot (\mat X - \mat A \mat S)\|_F^2 + \lambda \|\mat L \odot (\mat Y - \mat B \mat S)\|_F^2$\\
            $(\|\cdot\|_F,D(\cdot\|\cdot))$-SSNMF & $\argmin \limits_{\mat A, \mat B, \mat S\geq 0} \|\mat W \odot (\mat X - \mat A \mat S)\|_F^2 + \lambda D(\mat L\odot \mat Y \| \mat L \odot \mat B \mat S)$ \\
            $(D(\cdot\|\cdot),\|\cdot\|_F)$-SSNMF & $\argmin \limits_{\mat A, \mat B, \mat S\geq 0} D(\mat W\odot \mat X\| \mat W\odot \mat A \mat S) + \lambda \|\mat L\odot(\mat Y - \mat B \mat S)\|_F^2$ \\
            $(D(\cdot\|\cdot),D(\cdot\|\cdot))$-SSNMF & $\argmin \limits_{\mat A, \mat B, \mat S\geq 0} D(\mat W\odot \mat X\| \mat W\odot \mat A \mat S) + \lambda D(\mat L\odot \mat Y \| \mat L \odot \mat B \mat S)$ \\%
            \hline
\end{tabular}}}
\end{table}

\section{SSNMF Models: Motivation and Methods}
\label{sec: I-SSNMF method}
In this section, we present a statistical MLE motivation of several variants of the SSNMF model, introduce the general semi-supervised models, and provide a multiplicative updates method for each variant.
While historically the focus of SSNMF studies have been on classification~\cite{lee2009semi}, 
we highlight that this joint factorization model can be applied quite naturally to 
regression tasks. 

\subsection{Maximum Likelihood Estimation}\label{subsec:MLE}
In this section, we demonstrate that specific forms of our proposed variants of SSNMF are maximum likelihood estimators for given models of uncertainty or noise in the data matrices $\mat X$ and $\mat Y$.  These different uncertainty models have their likelihood function maximized by 
different error functions chosen for reconstruction and supervision errors, $R$ and $S$.  We summarize these results next; each MLE derived is a specific instance of a general model discussed in Section~\ref{subsec:propmodels} or in~\cite{lee2009semi}.

\begin{mle}
Suppose that the observed data $\mat X$ and supervision information $\mat Y$ have entries given as the sum of random variables, 
\begin{equation*}
    \mat X_{\gamma,\tau} = \sum_{i=1}^r x_{\gamma,i,\tau} \; \text{ and } \; \mat Y_{\eta,\tau} = \sum_{i=1}^r y_{\eta,i,\tau},
\end{equation*} 
and that the set of $\mat X_{\gamma,\tau}$ and $\mat Y_{\eta,\tau}$ are statistically independent conditional on $\mat A, \mat B$, and $\mat S$.
\begin{enumerate}
\itemsep-0.9em
\item When $x_{\gamma,i,\tau}$ and $y_{\eta,i,\tau}$ have distributions 
\begin{equation*}
    \mathcal{N}\pars{x_{\gamma,i,\tau} \middle| \mat A_{\gamma,i} \mat S_{i,\tau}, \sigma_1} \text{ and } \; \mathcal{N}\pars{y_{\eta,i,\tau} \middle| \mat B_{\eta,i} \mat S_{i,\tau}, \sigma_2}
\end{equation*} respectively, the maximum likelihood estimator is 
\begin{equation*}\label{fro-fro-MLE}
  \argmin_{\mat A, \mat B, \mat S \ge 0} \;\|\mat X - \mat A \mat S\|_F^2 + \frac{\sigma_1}{\sigma_2} \|\mat Y - \mat B \mat S\|_F^2.  
\end{equation*}

\item When $x_{\gamma,i,\tau}$ and $y_{\eta,i,\tau}$ have distributions $$\mathcal{N}\pars{x_{\gamma,i,\tau} \middle| \mat A_{\gamma,i} \mat S_{i,\tau}, \sigma_1} \text{ and } \mathcal{PO}\pars{y_{\eta,i,\tau} \middle| \mat B_{\eta,i} \mat S_{i,\tau}}$$ respectively, the maximum likelihood estimator is
$$\argmin_{\mat A, \mat B, \mat S \ge 0} \|\mat X - \mat A \mat S\|_F^2 +   2r\sigma_1 D(\mat Y\| \mat B \mat S).$$ \label{fro-div-MLE}

\item When $x_{\gamma,i,\tau}$ and $y_{\eta,i,\tau}$ have distributions $$\mathcal{PO}\pars{x_{\gamma,i,\tau} \middle| \mat A_{\gamma,i} \mat S_{i,\tau}} \text{ and } \mathcal{N}\pars{y_{\eta,i,\tau} \middle| \mat B_{\eta,i} \mat S_{i,\tau},\sigma_2}$$ respectively, the maximum likelihood estimator is 
$$\argmin_{\mat A, \mat B, \mat S \ge 0} D(\mat X\| \mat A \mat S) + \frac{1}{2r\sigma_2} \|\mat Y - \mat B \mat S\|_F^2.$$ \label{div-fro-MLE} 

\item When $x_{\gamma,i,\tau}$ and $y_{\eta,i,\tau}$ have distributions $$x_{\gamma,i,\tau} \sim \mathcal{PO}\pars{x_{\gamma,i,\tau} \middle| \mat A_{\gamma,i} \mat S_{i,\tau}} \text{ and } \mathcal{PO}\pars{y_{\eta,i,\tau} \middle| \mat B_{\eta,i} \mat S_{i,\tau}}$$ respectively, the maximum likelihood estimator is 
$$\argmin_{\mat A, \mat B, \mat S \ge 0} D(\mat X\| \mat A \mat S) + D(\mat Y\| \mat B \mat S).$$ \label{div-div-MLE}
\end{enumerate}
\end{mle}
\vspace{-0.5cm}

We note that \ref{div-div-MLE} follows from \cite{cemgil2008bayesian,favaro20073,virtanen2008bayesian}, but the others are distinct from previous MLE derivations due to the difference in the 
distributions assumed on data $\mat X$ and supervision $\mat Y$.  
Here, we provide only the MLE derivation for \ref{fro-div-MLE} as the other derivations are similar; these are included in the appendix for completeness.
We demonstrate that the MLE, in the case that the uncertainty on $\mat X$ is Gaussian distributed and on $\mat Y$ is Poisson distributed, is a specific instance of the $(\|\cdot\|_F,D(\cdot\|\cdot))$-SSNMF
model. 

Our models for the distribution of observed entries of $\mat X$ and $\mat Y$ assume that the mean is given by  
$\mathbb{E}[\mat X] = \mat A \mat S$ and $\mathbb{E}[\mat Y] = \mat B \mat S$, and the uncertainty in the set of observations in $\mat X$ is governed by a Gaussian distribution while the set in $\mat Y$ is governed by a Poisson distribution.
That is, we consider hierarchical models for $\mat X$ and $\mat Y$ where $$\mat X_{\gamma,\tau} = \sum_{i=1}^r x_{\gamma,i,\tau} \text{ and } x_{\gamma,i,\tau} \sim \mathcal{N}\pars{x_{\gamma,i,\tau} \middle| \mat A_{\gamma,i} \mat S_{i,\tau}, \sigma_1},$$ 
$$
\mat Y_{\eta,\tau} = \sum_{i=1}^r y_{\eta,i,\tau} \text{ and } y_{\eta,i,\tau} \sim \mathcal{PO}\pars{y_{\eta,i,\tau} \middle| \mat B_{\eta,i} \mat S_{i,\tau}}.$$
Note then that 
{\footnotesize \begin{equation*}\mat X_{\gamma,\tau} \sim \mathcal{N}\pars{\mat X_{\gamma,\tau} \middle| \sum_{i=1}^r \mat A_{\gamma,i} \mat S_{i,\tau}, r\sigma_1}, \text{ and }
\end{equation*}
\begin{equation*}
\mat Y_{\eta,\tau} \sim \mathcal{PO}\pars{\mat Y_{\eta,\tau} \middle| \sum_{i=1}^r \mat B_{\eta,i} \mat S_{i,\tau}}
\end{equation*}}%
due to the summable property of Gaussian and Poisson random variables.  We note that this assumes different distributions on the two collections of rows of the NMF \eqref{eq:stackedNMF}, with Gaussian and Poisson models of uncertainty. 

Assuming that the set of $\mat X_{\gamma,\tau}$ and $\mat Y_{\eta,\tau}$ are statistically independent conditional on $\mat A$, $\mat B$, and $\mat S$, 
we have that the likelihood $p(\mat X, \mat Y| \mat A, \mat B, \mat S)$ is 

{\footnotesize%
\begin{equation}
    \prod_{\gamma,\tau} \mathcal{N}\pars{\mat X_{\gamma,\tau} \middle| \sum_{i=1}^r \mat A_{\gamma,i} \mat S_{i,\tau},   r\sigma_1   } \prod_{\eta,\tau} \mathcal{PO}\pars{  \mat Y_{\eta,\tau} \middle| \sum_{i=1}^r \mat B_{\eta,i} \mat S_{i,\tau}}. \label{eq:frodivlikelihood}
\end{equation}}

\noindent We apply the monotonic natural logarithmic function to the likelihood and ignore terms that are invariant with the factor matrices.  This transforms the likelihood into a $(\|\cdot\|_F,D(\cdot\|\cdot))$-SSNMF objective while preserving the maximizer.  That is, the log likelihood (excluding additive terms which are constant with respect to $\mat A$, $\mat B$, and $\mat S$) is 

{\footnotesize%
\begin{align*}
&\ln p\pars{\mat X, \mat Y \middle| \mat A, \mat B, \mat S} =^+ -\frac{1}{2r\sigma_1} \sum_{\gamma,\tau}\pars{\mat X_{\gamma,\tau} - \sum_{i=1}^r \mat A_{\gamma,i} \mat S_{i,\tau}}^2 
\\&\hspace{1.1cm}- \sum_{\eta,\tau} \bigg[(\mat B\mat S)_{\eta,\tau} -   \mat Y_{\eta,\tau} \log(\mat B\mat S)_{\eta,\tau} 
+ \log\Gamma\pars{    \mat Y_{\eta,\tau} + 1}\bigg]
\\&\quad\quad\quad=^+ -\frac{1}{2r\sigma_1}\bracs{     \|\mat X - \mat A\mat S\|_F^2 +   2r\sigma_1 D(\mat Y\|  \mat B\mat S)}.
\end{align*}}%

\noindent Here, $=^+$ denotes suppression of additive terms that do not depend upon $\mat A$, $\mat B$, and $\mat S$.  Thus, the maximum likelihood estimators for $\mat A$, $\mat B$, and $\mat S$ are given by  
\begin{equation*}
\argmin_{\mat A,\mat B,\mat S \ge 0} \|\mat X - \mat A\mat S\|_F^2 +   2r\sigma_1 D(\mat Y\| \mat B\mat S).    
\end{equation*}
We see that the MLE in the case of Gaussian uncertainty on the observations in $\mat X$ and Poisson uncertainty on the observations in $\mat Y$, is a specific instance of the $(\|\cdot\|_F,D(\cdot\|\cdot))$-SSNMF objective where the regularization parameter $\lambda$ is a multiple of the variance of the Gaussian distribution.
The other MLEs are derived similarly; see Appendix~\ref{sec:appendixA}.  

An instance of each of the models in Table~\ref{table:models} are MLE for a given model of uncertainty in the observed data $\mat X$ and supervision $\mat Y$.  While this motivates our exploration of these models, we present them in more general context next and provide training methods for the general form.

\subsection{General Models and Mult.\ Updates}\label{subsec:propmodels}
Recall that $(\|\cdot\|_F,\|\cdot\|_F)$-SSNMF
is defined by \eqref{eqn:ssnmf} and multiplicative updates are derived in~\cite{lee2009semi}.  Now, we propose the general form of $(\|\cdot\|_F,D(\cdot\|\cdot))$-SSNMF, $(D(\cdot\|\cdot),\|\cdot\|_F)$-SSNMF, and $(D(\cdot\|\cdot),D(\cdot\|\cdot))$-SSNMF and present multiplicative updates methods for each model. 
These three models are novel forms of SSNMF, and besides their statistical motivation via MLE, we demonstrate their promise experimentally in Section \ref{sec: numerical experiments}.

As in \cite{lee2009semi}, our multiplicative updates methods allow for missing (or certainty-weighted) data and missing (or certainty-weighted) supervision information via matrices $\mat W$ and $\mat L$, which represent our knowledge or certainty of the corresponding entries of $\mat X$ and $\mat Y$, respectively.  When $\mat W$ is a matrix of all ones (or more generally has all equal entries) and $\mat L$ is a matrix of all zeros, the SSNMF models reduce to either the $\|\cdot\|_F$-NMF or $D(\cdot\|\cdot)$-NMF. 
The SSNMF model is fully supervised when $\text{supp}(\mat Y) \subset \text{supp}(\mat L)$ and $\mat Y$ contains supervision information for each element in $\mat X$.

The first proposed semi-supervised NMF model is $(\|\cdot\|_F,D(\cdot\|\cdot))$-SSNMF, which is defined by the solution to 
\begin{equation}
\label{eq:FI-SSNMF}
  \argmin_{\mat A,\mat B,\mat S \ge 0} \|\mat W \odot (\mat X - \mat A\mat S)\|_F^2 + \lambda D(\mat L \odot \mat Y \| \mat L \odot \mat B\mat S).   
\end{equation}
We denote this objective function as $F_2(\mat A, \mat B, \mat S; \mat X, \mat Y)$.  Similar to the previous SSNMF model, this model seeks a joint factorization of the data matrix $\mat X$ and target matrix $\mat Y$; however, the error functions applied to the reconstruction and classification terms in the objective differ.

The multiplicative updates for this model are provided in Algorithm~\ref{algo:fissnmf}.  We provide intuition for the derivation of only this method as the others are similar. 
The multiplicative updates for $\mat A$, $\mat B$, and $\mat S$ which minimize~\eqref{eq:FI-SSNMF} are derived as follows.
The gradient of the objective function of~\eqref{eq:FI-SSNMF} with respect to $\mat A$, $\mat B$ and $\mat S$ are, respectively,
\begin{align*}
\nabla_{\mat A} F_2 &= -2[\mat W \odot (\mat X-\mat A\mat S)]\mat S^\top,\\
\nabla_{\mat B} F_2 &= \mat L\mat S^\top - \left [ \frac{\mat L \odot \mat Y}{\mat L \odot \mat B\mat S} \odot \mat L \right ]\mat S^\top, \text{ and }\\
\nabla_{\mat S} F_2 &= \lambda  \mat B^\top \mat L - \lambda \mat B^\top \hspace{-1mm}\left[\frac{\mat L \odot \mat Y}{\mat L \odot \mat B\mat S} \odot \mat L  \right] -2\mat A^\top [\mat W \hspace{-1mm}\odot(\mat X\hspace{-1mm}-\hspace{-1mm}\mat A\mat S)].
\end{align*}%
The multiplicative updates method, Algorithm~\ref{algo:fissnmf}, can be viewed as an entrywise gradient descent method, where the stepsizes are chosen individually for each entry of the updating matrix to ensure nonnegativity. That is, the updates in Algorithm~\ref{algo:fissnmf} are given by 
\begin{align*}
\mat A &\rightarrow \mat A - \Gamma \odot \nabla_{\mat A} F_2 \text{ \;\;when\;\; } \Gamma = \frac{\mat A}{2(\mat W\odot \mat A\mat S)\mat S^\top},\\
\mat B &\rightarrow \mat B - \Gamma \odot \nabla_{\mat B} F_2 \text{ \;\;when\;\; } \Gamma = \frac{\mat B}{\mat L\mat S^\top}, \text{ and }\\
\mat S &\rightarrow \mat S - \Gamma \odot \nabla_{\mat S} F_2 \text{ \;\;when\;\; } \Gamma = \frac{\mat S}{2\mat A^\top(\mat W \odot \mat A\mat S) + \lambda \mat B^\top \mat L}.
\end{align*}

The next proposed semi-supervised NMF model is $(D(\cdot\|\cdot),\|\cdot\|_F)$-SSNMF, 
defined by the solution to 
\begin{equation}
\label{eq:IF-SSNMF}
  \argmin_{\mat A,\mat B,\mat S \ge 0} D(\mat W \odot \mat X \| \mat W \odot \mat A\mat S) + \lambda \|\mat L \odot (\mat Y - \mat B\mat S)\|_F^2.   
\end{equation}
We denote this objective function as $F_3(\mat A, \mat B, \mat S; \mat X, \mat Y)$.
Again, this model seeks a joint factorization of the data matrix $\mat X$ and target matrix $\mat Y$; here the reconstruction error is penalized by the information divergence, while the supervision error is penalized by the Frobenius norm.  
Multiplicative updates for this model are provided in Algorithm~\ref{algo:ifssnmf}.

The third, and final, proposed semi-supervised NMF model is $(D(\cdot\|\cdot),D(\cdot\|\cdot))$-SSNMF, 
defined by the solution to 
\begin{equation}
\label{eq:II-SSNMF}
  \argmin_{\mat A,\mat B,\mat S \ge 0} D(\mat W \odot \mat X \| \mat W \odot \mat A\mat S) + \lambda D(\mat L \odot \mat Y \| \mat L \odot \mat B\mat S).   
\end{equation}
We denote this objective function as $F_4(\mat A, \mat B, \mat S; \mat X, \mat Y)$.
Again, this model seeks a joint factorization of the data matrix $\mat X$ and target matrix $\mat Y$; here both the reconstruction error and supervision error are penalized by the information 
divergence error function.  
The multiplicative updates for this model are provided in Algorithm~\ref{algo:iissnmf}.

\begin{algorithm}[t]
    \setstretch{1.5}
    \caption{$(\|\cdot\|_F,D(\cdot\|\cdot))$-SSNMF mult. updates}
    \label{algo:fissnmf}
    \begin{algorithmic}[1]
    \REQUIRE{$\mat X, \mat W \in \mathbb{R}^{n_1 \times n_2}_{\ge 0}$, $\mat Y, \mat L \in \mathbb{R}^{k \times n_2}_{\ge 0}$, $r$, $\lambda$, $N$}
    \STATE{Initialize $\mat A \in \mathbb{R}^{n_1 \times r}_{\ge 0}, \mat S \in \mathbb{R}^{r \times n_2}_{\ge 0}, \mat B \in \mathbb{R}^{k \times r}_{\ge 0} $}
    \FOR{$i = 1, ..., N$}
        \STATE {$\mat A \leftarrow \mat A \odot \frac{(\mat W \odot \mat X) \mat S^\top}{(\mat W \odot \mat A\mat S) \mat S^\top}$\;}
        \STATE {$\mat B \leftarrow \frac{\mat B}{ \mat L  \mat S ^\top} \odot \left[\frac{(\mat L \odot \mat Y)}{ (\mat L \odot \mat B\mat S)} \odot \mat L\right] \mat S^\top $\;}
        \STATE {$\mat S \leftarrow \mat S \odot \frac{ 2 \mat A^\top (\mat W \odot \mat X) + \lambda \mat B^\top \left[\frac{(\mat L \odot \mat Y)}{(\mat L \odot \mat B\mat S)} \odot \mat L \right]}{2\mat A^\top (\mat W \odot \mat A \mat S) + \lambda \mat B^\top \mat L }  $\;}
        \ENDFOR
    \end{algorithmic}
\end{algorithm}
\begin{algorithm}[t]
    \setstretch{1.5}
    \caption{$(D(\cdot\|\cdot),\|\cdot\|_F)$-SSNMF mult. updates}
    \label{algo:ifssnmf}
    \begin{algorithmic}[1]
    \REQUIRE{$\mat X, \mat W \in \mathbb{R}^{n_1 \times n_2}_{\ge 0}$, $\mat Y, \mat L \in \mathbb{R}^{k \times n_2}_{\ge 0}$, $r$, $\lambda$, $N$}
    \STATE{Initialize $\mat A \in \mathbb{R}^{n_1 \times r}_{\ge 0}, \mat S \in \mathbb{R}^{r \times n_2}_{\ge 0}, \mat B \in \mathbb{R}^{k \times r}_{\ge 0} $}
    \FOR{$i = 1, ..., N$}
        \STATE {$\mat A \leftarrow \frac{\mat A}{\mat W \mat S^\top} \odot \left[\frac{(\mat W \odot \mat X)}{(\mat W \odot \mat A\mat S)} \odot \mat W\right] \mat S^\top$\;}
        \STATE {$\mat B \leftarrow \mat B \odot \frac{(\mat L \odot \mat Y)\mat S^\top}{(\mat L \odot \mat B\mat S)\mat S^\top}$\;}
        \STATE {$\mat S \leftarrow \mat S \odot  \frac{\mat A^\top \left[\frac{(\mat W \odot \mat X)}{(\mat W \odot \mat A\mat S)} \odot \mat W\right] + 2\lambda \mat B^\top (\mat L \odot \mat Y)}{\mat A^\top \mat W + 2\lambda \mat B^\top (\mat L \odot \mat B\mat S)}$\;}
        \ENDFOR
    \end{algorithmic}
\end{algorithm}
\begin{algorithm}[ht]
    \setstretch{1.5}
    \caption{$(D(\cdot\|\cdot),D(\cdot\|\cdot))$-SSNMF mult. updates}
    \label{algo:iissnmf}
    \begin{algorithmic}[1]
    \REQUIRE{$\mat X, \mat W \in \mathbb{R}^{n_1 \times n_2}_{\ge 0}$, $\mat Y, \mat L \in \mathbb{R}^{k \times n_2}_{\ge 0}$, $r$, $\lambda$, $N$}
    \STATE{Initialize $\mat A \in \mathbb{R}^{n_1 \times r}_{\ge 0}, \mat S \in \mathbb{R}^{r \times n_2}_{\ge 0}, \mat B \in \mathbb{R}^{k \times r}_{\ge 0} $}
    \FOR{$i = 1, ..., N$}
        \STATE {$\mat A \leftarrow \frac{\mat A}{\mat W \mat S^\top} \odot \left[\frac{(\mat W \odot \mat X)}{(\mat W \odot \mat A\mat S)} \odot \mat W\right] \mat S^\top$\;}
        \STATE {$\mat B \leftarrow \frac{\mat B}{\mat L\mat S^\top} \odot \left[\frac{(\mat L \odot \mat Y)}{(\mat L \odot \mat B\mat S)} \odot \mat L\right] \mat S^\top$\;}
        \STATE {$\mat S \leftarrow \mat S \odot \frac{\mat A^\top \left[\frac{(\mat W \odot \mat X)}{(\mat W \odot \mat A\mat S)} \odot \mat W\right] + \lambda \mat B^\top \left[\frac{(\mat L \odot \mat Y)}{(\mat L \odot \mat B\mat S)} \odot \mat L\right]}{\mat A^\top \mat W + \lambda \mat B^\top \mat L}$\;}
        \ENDFOR
    \end{algorithmic}
\end{algorithm}

As previously stated in Section~\ref{subsec:MLE}, an instance of each family of models, $(\|\cdot\|_F,\|\cdot\|_F)$-SSNMF, $(\|\cdot\|_F,D(\cdot\|\cdot))$-SSNMF, $(D(\cdot\|\cdot),\|\cdot\|_F)$-SSNMF, and $(D(\cdot\|\cdot),D(\cdot\|\cdot))$-SSNMF, correspond to the MLE in the case that the data $\mat X$ and supervision $\mat Y$ are sampled from specific distributions with mean given by a latent lower-dimensional factorization model. One might expect that each model is most appropriately applied when the associated model of uncertainty is a reasonable assumption (i.e., one has \emph{a priori} information indicating so).
For example, we expect that the Gaussian uncertainty assumption associated to $(\|\cdot\|_F,\|\cdot\|_F)$-SSNMF is likely most appropriate when the supervised learning task is a regression task, and likely not as appropriate for a classification task where the targets have discrete form.

We note that the iteration complexity of each of these methods scales with complexity of 
multiplication of matrices of size $n_1 \times \max\{k,r\}$ and $\max\{k,r\} \times n_2$.  In our implementation of each of these methods, we ensure that there is no division by zero by adding a small positive value to all entries of divisors. Implementation of these methods and code for experiments is available in the Python package \texttt{SSNMF}~\cite{SSNMFpackage}.  
Finally, we note that the behavior of these models and methods are highly dependent on the hyperparameters $r$, $\lambda$, and $N$.  One can select the parameters according to \emph{a priori} information or use a heuristic 
selection technique; 
we use both and indicate selected parameters and method of selection.

\subsection{Classification Framework} \label{sec:classification}
Here we describe a framework for using any of the SSNMF models for classification tasks.  
Given training data $\mat X_{\text{train}}$ (with any missing data indicated by matrix $\mat W_{\text{train}}$) and labels $\mat Y_{\text{train}}$, and testing data $\mat X_{\text{test}}$ (with unknown data indicated by matrix $\mat W_{\text{test}}$), 
we first train our $(R(\cdot \| \cdot),S(\cdot \| \cdot))$-SSNMF model 
to obtain learned dictionaries $\mat A_{\text{train}}$ and $\mat B_{\text{train}}$. We then use these learned matrices to obtain the representation of test data in the subspace spanned by $\mat A_{\text{train}}$, $\mat S_{\text{test}}$, and the predicted labels for the test data $\mat Y_{\text{test}}$.  This process is: 
\begin{enumerate}
    \item Compute $\mat A_{\text{train}}, \mat B_{\text{train}}, \mat S_{\text{train}}$ as \[ \argmin\limits_{\mat A, \mat B, \mat S\geq 0} R(\mat W_{\text{train}} \odot \mat X_{\text{train}}, \mat W_{\text{train}} \odot \mat A\mat S) + \lambda S(\mat Y_{\text{train}}, \mat B\mat S).\]
    \item Solve $ \mat S_{\text{test}} = \argmin\limits_{\mat S\geq 0} R(\mat W_{\text{test}} \odot \mat X_{\text{test}}, \mat W_{\text{test}} \odot \mat A_{\text{train}}\mat S).$
    \item 
    Compute predicted labels as $\hat{\mat Y}_{\text{test}} = \text{label}(\mat B_{\text{train}}\mat S_{\text{test}})$, where $\text{label}(\cdot)$ assigns the largest entry of each column to 1 and all other entries to 0.
\end{enumerate}
In step 1, we compute $\mat A_{\text{train}}, \mat B_{\text{train}}$, and $\mat S_{\text{train}}$ using implementations of the multiplicative updates methods described above.  In step 2, we use either a nonnegative least-squares method (if $R = \|\cdot\|_F$) or one-sided multiplicative updates only updating $\mat S_{\text{test}}$ (if $R = D(\cdot \| \cdot)$). We note that this framework is significantly different than the classification framework proposed in~\cite{lee2009semi}; in particular, we use the classifier $\mat B$ learned by SSNMF, rather than independent SVM trained on the SSNMF-learned lower-dimensional 
representation.

\section{Numerical Experiments}
\label{sec: numerical experiments}
In this section, we 
present numerical experiments of the proposed models applied to both synthetic and real data.

\subsection{Synthetic Data Experiments}\label{subsec:syntheticdata}

We expect that, for each pair of distributions of uncertainty, the MLE model derived in Section~\ref{subsec:MLE} will produce larger likelihood (i.e., smaller relative error) than any other SSNMF model. To confirm this hypothesis, we generate synthetic data according to the four distributions in Section~\ref{subsec:MLE}, train the SSNMF models, and measure the relative error (error is negative likelihood).  

Through all experiments, we use the same factor matrices $\mat A \in \mathbb{R}^{500 \times 5}, \mat S \in \mathbb{R}^{5 \times 500},$ and $\mat B \in \mathbb{R}^{500 \times 5}$.  Here $\mat A$ is generated with all entries sampled from the uniform distribution on $[0,1]$, and $\mat S$ and $\mat B$ are generated as sparse random matrices with density $0.5$ (support sampled uniformly amongst matrix entries) with nonzero entries sampled from the uniform distribution on $[0,1]$. In each experiment, we train each of the SSNMF models with rank $r = 5$ and $\lambda = 1$ by running $N = 100000$ iterations of the multiplicative updates of~\cite{lee2009semi} and Algorithms~\ref{algo:fissnmf}, \ref{algo:ifssnmf}, and \ref{algo:iissnmf} using all data and supervision; each algorithm is initialized with the same matrices $\mat A^{(0)}, \mat B^{(0)},$ and $\mat S^{(0)}$ and we denote the resulting approximate factor matrices as $\mat A^{(N)}, \mat B^{(N)},$ and $\mat S^{(N)}$. We present relative errors averaged over five trials (independent initializations of $\mat A^{(0)}, \mat B^{(0)},$ and $\mat S^{(0)}$).  Here, we let the relative error for objective function $F$ be 
\begin{equation}
    \frac{F(\mat A^{(N)}, \mat B^{(N)}, \mat S^{(N)}; \mat A \mat S, \mat B \mat S)}{F(\mat A^{(0)}, \mat B^{(0)}, \mat S^{(0)}; \mat A \mat S, \mat B \mat S)}. \label{eq:relative error}
\end{equation}

In our first experiment, we generate data $\mat X_{ij} \sim \mathcal{N}\pars{\mat X_{ij} \middle| (\mat A\mat S)_{ij},1}$ and supervision $\mat Y_{ij} \sim \mathcal{N}\pars{\mat Y_{ij} \middle| (\mat B\mat S)_{ij},1}$, 
and compare each model's average relative error for objective function $F_1$.
In our second experiment, we generate data $\mat X_{ij} \sim \mathcal{N}\pars{\mat X_{ij} \middle| (\mat A\mat S)_{ij},1/2r}$ and supervision $\mat Y_{ij} \sim \mathcal{PO}\pars{\mat Y_{ij} \middle| (\mat B\mat S)_{ij}}$, 
and compare each model's average relative error for objective function $F_2$.
In our third experiment, we generate data $\mat X_{ij} \sim \mathcal{PO}\pars{\mat X_{ij} \middle| (\mat A\mat S)_{ij}}$ and supervision $\mat Y_{ij} \sim \mathcal{N}\pars{\mat Y_{ij} \middle| (\mat B\mat S)_{ij},1/2r}$, 
and compare each model's average relative error for objective function $F_3$.
In our last experiment, we generate data $\mat X_{ij} \sim \mathcal{PO}\pars{\mat X_{ij} \middle| (\mat A\mat S)_{ij}}$ and supervision $\mat Y_{ij} \sim \mathcal{PO}\pars{\mat Y_{ij} \middle| (\mat B\mat S)_{ij}}$, 
and compare each model's average relative error for objective function $F_4$.

All results are given in Table~\ref{table:syntheticexperiment}; lowest average relative errors are listed in bold for all experiments.  Note that in each experiment, as expected, the MLE model produces smaller relative error than other SSNMF models.

\begin{table}[tb] 
\caption{Experimental results with synthetic data.}\label{table:syntheticexperiment}
{ \renewcommand{\arraystretch}{1.15}
\footnotesize
\centering
\begin{tabular}{| c | c | c | c | c |}
            \hline
            Experiment & 1 & 2 & 3 & 4 \\
            \hline
      SSNMF & $F_1$ err.\ & $F_2$ err.\ & $F_3$ err.\ & $F_4$ err.\ \\
            \hline
            $(\|\cdot\|_F,\|\cdot\|_F)$ & \textbf{0.2917} & 0.0081 & 0.0078 & 0.0204 \\
            \hline
            $(\|\cdot\|_F,D(\cdot\|\cdot))$ & 0.2927 & \textbf{0.0065} & 0.0109 & 0.0199 \\
            \hline
            $(D(\cdot\|\cdot),\|\cdot\|_F)$ & 0.2930 & 0.0093 & \textbf{0.0071} & 0.0210 \\
            \hline
            $(D(\cdot\|\cdot),D(\cdot\|\cdot))$ & 0.2922 & 0.0076 & 0.0075  & \textbf{0.0172} \\
\hline
\end{tabular}}
\end{table}

\subsection{20 Newsgroups Data Experiments}\label{subsec:20newsdata}
    The 20 Newsgroups data set~\cite{20news} is a collection of approximately 20,000 newsgroup documents.\footnote{Our results present this data in its raw form; in particular, we do not capitalize words to reflect common usage. Results are in no way meant to be a political statement.}
The data set consists of six groups partitioned roughly according to subjects, with a total of 20 subgroups, and is commonly used as an experimental benchmark for 
document classification and 
clustering; see e.g.,~\cite{lee2009semi}. Here, we compare classification accuracy, unlike in Section~\ref{subsec:syntheticdata} where we compare reconstruction errors.

We consider a subset of the data set, summarized in Table~\ref{table:20news_data}. We treat the groups as classes and assign them labels, and we treat the subgroups as (un-labeled) latent topics in the data.  We remove headers, footers, and quotes from all documents, subsample the data set to obtain a balanced data set across classes (1796 document per class), and split the data set into train (60\%), validation (20\%), and test (20\%) sets.
We compute the term frequency–inverse document frequency (TF-IDF) representation for documents
using TFIDFVectorizer~\cite{scikit-learn}.
The NLTK English stopword list~\cite{nltk}, and words appearing in less than 5 documents, or more than 70\% of the documents were removed.
We use the tokenizer \texttt{[a-zA-Z]+}
and limit the vocabulary size to 5000.
We 
compare to the linear Support Vector Machine (SVM) classifier and Multinomial Naive Bayes (NB) (see e.g.,~\cite{manning2008introduction}) using the Scikit-learn implementation with default parameters~\cite{scikit-learn}, where the groups in Table~\ref{table:20news_data} are treated as classes.
We consider all SSNMF models with the training process described in Section~\ref{sec:classification} with the maximum number of iterations (number of multiplicative updates) $N = 50$; our stopping criterion is the earlier of $N$ iterations or relative error~\eqref{eq:relative error} below tolerance $tol$. 

\begin{table}[tb]
\centering
\footnotesize

\caption{20 Newsgroups and subgroups.}\label{table:20news_data}
{ \renewcommand{\arraystretch}{1.15}
\begin{tabular}{ l  l }
            \hline
            Groups &\qquad Subgroups \\
            \hline
            Computers &\qquad graphics, mac.hardware, windows.x\\
            Sciences &\qquad crypt(ography), electronics, space\\
            Politics &\qquad guns, mideast \\
            Religion &\qquad atheism, christian(ity)\\
            Recreation &\qquad autos, baseball, hockey\\
\hline
\end{tabular}}
\end{table}

We also apply SVM as a classifier to the low-dimensional representation obtained from NMF as follows.
We consider the default implementation~\cite{scikit-learn} of $\|\cdot\|_F$-NMF  with multiplicative updates, random initialization, and maximum number of iterations $N = 400$. 
We apply NMF on the train data to obtain a vocabulary dictionary matrix $\mat A_{\text{train}}$ and a document representation $\mat S_{\text{train}}$.
Next, we train an SVM classifier using $\mat S_{\text{train}}$ and the labels of the train set.
We test our model by (i) computing the document representation of the test data $\mat S_{\text{test}}$ from the learned dictionary $\mat A_{\text{train}}$ (i.e., step 2 of Section~\ref{sec:classification}), then (ii) applying the trained SVM classifier on $\mat S_{\text{test}}$ to obtain the test predicted labels.

For both NMF and all four SSNMF models, we consider rank (the number of topics\footnote{A larger choice of rank could be made to learn hidden topics within subgroups.}) equal to 13.
We select the hyperparameters $tol$ and $\lambda$ for the models by searching over different values and selecting those with the highest average classification accuracy on the validation set; see Appendix~\ref{ap:20news}. 

\begin{table}[htb]
\centering
\caption{Mean (and std. dev.) of test classification accuracy for each of the models on 20 Newsgroups data.}\label{table:class_accuracy}
{ \renewcommand{\arraystretch}{1.15}
\footnotesize
\begin{tabular}{ |l|c| } 
\hline
Model & Class. accuracy \% (sd) \\
\hline
$(\|\cdot\|_F,\|\cdot\|_F)$ & 79.37 (0.47)  \\ 
$(\|\cdot\|_F,D(\cdot\|\cdot))$  &79.51 (0.38)  \\ 
$(D(\cdot\|\cdot),\|\cdot\|_F)$ & \textbf{81.88} (0.44)  \\ 
$(D(\cdot\|\cdot),D(\cdot\|\cdot))$ & 81.50 (0.47)  \\
$\|\cdot\|_F$-NMF + SVM & 70.99 (2.71)  \\
\hline
SVM &  80.70 (0.27)  \\
Multinomial NB  &  \textbf{82.28} \\
\hline
\end{tabular}}
\end{table}

We report in Table~\ref{table:class_accuracy} the mean and standard deviation of the test classification accuracy for each of the models over 11 trials.
We define the test classification accuracy as $\sum_{i=1}^{n} \delta(\mat Y_i, \hat{\mat Y}_i)/n$, where $\delta(u,v)= 1$ for $u=v$, and 0 otherwise, and where $\mat Y_i$ and $\hat{\mat Y}_i$ are true and predicted labels, respectively.
We observe that $(D(\cdot\|\cdot),\|\cdot\|_F)$-SSNMF produces the highest average classification accuracy, and is comparable to 
Multinomial NB.

In Table~\ref{table:class_accuracy}, we separate models that simultaneously perform dimensionality-reduction and classification from those which only perform classification.  Note that the SSNMF models, which provide both dimensionality-reduction and classification in that lower-dimensional space, do not suffer great accuracy loss over models which perform classification in the high-dimensional space.
We observe that the $(D(\cdot\|\cdot),\|\cdot\|_F)$-SSNMF performs significantly better than the $\|\cdot\|_F$-NMF + SVM in terms of accuracy.
In Appendix~\ref{ap:20news}, we present NMF and SSNMF model results, and compare keywords, classifier matrices, and clustering performance.
\begin{figure}[tb]
    \centering
    \includegraphics[width=\columnwidth]{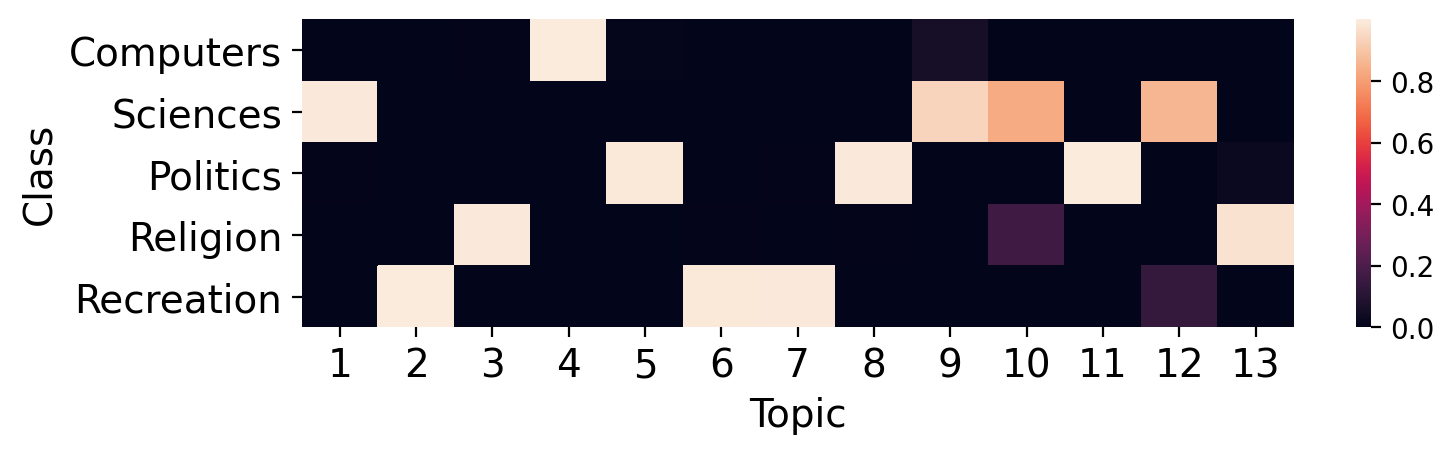}
    \caption{The normalized $\mat B_{\text{train}}$ matrix for the $(D(\cdot\|\cdot),\|\cdot\|_F)$ SSNMF decomposition corresponding to the median test classification accuracy equal to 81.78.}
    \label{fig:SSNMF_Model5}
\end{figure}

Here, we consider the ``typical" decomposition for the $(D(\cdot\|\cdot),\|\cdot\|_F)$-SSNMF by selecting the decomposition corresponding to the median test classification accuracy.
We display in Figure~\ref{fig:SSNMF_Model5} the column-sum normalized $\mat B_{\text{train}}$ matrix of the decomposition, where each column illustrates the distribution of topic association to classes.
We display in Table~\ref{table:model5_keywords} the top 10 keywords (i.e. those that have the highest weight in topic column of $\mat A_{\text{train}}$) for each topic of the $(D(\cdot\|\cdot),\|\cdot\|_F)$-SSNMF of Figure~\ref{fig:SSNMF_Model5}.

We (qualitatively) observe from Table~\ref{table:model5_keywords} that topic 5 (``armenian", ``fbi"), topic 8 (``arab"), and topic 11 (``weapons") captures the subjects of Middle East and guns (``israel", ``government", ``gun"). 
All three topics are associated with class Politics; see Figure~\ref{fig:SSNMF_Model5}.
We also observe that topic 1 (``space", ``government") and topic 9 (``chip", ``key", ``algorithm") relate to electronics/cryptography. Both are associated to class Sciences.
Topic 2 is related to autos (``car", ``engine"), topic 6 captures the specific subject of baseball, and topic 7 of hockey.
Indeed, all three topics are associated to class Recreation, and topic 12 (``available",``key",``phone") is shared between Sciences and Recreation.
Topics 3 and 13 capture topics related to religion and beliefs (``god", ``believe", ``religious") and are associated to class Religion.
Topic 10 (``earth", ``space") is shared between Religion and Sciences.
Topic 4 captures computer subjects (``x" for Windows 10, ``graphics", and ``mac"). Indeed, topic 4 is the only topic associated to class Computers in Figure~\ref{fig:SSNMF_Model5}. 

While the learned topics in Table~\ref{table:model5_keywords} are not one-to-one with the subgroups in Table~\ref{table:20news_data}, these topics appear relatively coherent.  We see in Table~\ref{table:class_accuracy} that these learned topics serve the classification task well; that is, the data representation in this significantly lower-dimensional space is able to achieve nearly the same accuracy as the higher-dimensional multinomial NB model.  We expect that this is due to the relevance of the topic modeling and classification tasks on this data set.  We note that while $(D(\cdot\|\cdot),\|\cdot\|_F)$-SSNMF outperforms the other SSNMF models in terms of \textit{classification accuracy}, this does not imply that a different model would not produce a lower overall \emph{relative error} for the objective function.  Additionally, the strong performance of $(D(\cdot\|\cdot),\|\cdot\|_F)$-SSNMF could be due to a number of factors including the choice of hyperparameters. Ongoing and future work will further investigate this phenomenon.

\begin{table*}[htb]
 \resizebox{\textwidth}{!}{
    \centering
\begin{tabular}{ |c|c|c|c|c|c|c|c|c|c|c|c|c|} 
\hline
Topic 1 & Topic 2 & Topic 3 & Topic 4 & Topic 5 & Topic 6 & Topic 7 & Topic 8 & Topic 9 & Topic 10 & Topic 11 & Topic 12 & Topic 13 \\
\hline
would & game & god & x & would & game & players & people & would & one & israel & like & god \\
space & team & would & thanks & armenian & one & team & israel & chip & us & guns & anyone & people \\
government & car & one & anyone & one & like & car & gun & key & get & people & available & church \\
use & games & jesus & graphics & people & car & last & right & algorithm & could & gun & key & one \\
key & engine & think & know & fbi & baseball & year & government & use & like & well & probably & christians \\
chip & year & bible & use & armenians & think & game & us & using & earth & weapons & right & jesus \\
get & like & believe & mac & israeli & get & hockey & say & bit & space & know & phone & would \\
clipper & know & christian & please & killed & season & would & jews & like & know & like & another & religious \\
one & espn & christ & would & fire & last & go & arab & system & see & government & also & christian \\
could & get & say & get & jews & would & time & one & data & used & would & big & different \\
\hline
\end{tabular}
}
\caption{Top keywords representing each topic of the $(D(\cdot\|\cdot),\|\cdot\|_F)$-SSNMF model referred to in Figure~\ref{fig:SSNMF_Model5}.}
\label{table:model5_keywords}
\end{table*}

\section{Conclusion} \label{sec:conclusion}

In this work, we have have proposed several SSNMF models, and have demonstrated that these models and that of~\cite{lee2009semi} are MLE in the case of specific distributions of uncertainty assumed on the data and labels.  We provided multiplicative update training methods for each model, and demonstrated the ability of these models to perform classification.

In future work, we plan to take a Bayesian approach to SSNMF by assuming data-appropriate priors and performing maximum \emph{a posteriori} estimation.  Furthermore, we will form a general framework of MLE models for exponential family distributions of uncertainty, and study the class of models where multiplicative update methods are feasible.

\section*{Acknowledgements}
The authors are appreciative of useful conversations with William Swartworth, Joshua Vendrow, and Liza Rebrova.  

\bibliographystyle{plain}
\bibliography{main}

\newpage
\section{Appendix}
\appendix
In this appendix, we provide the remaining MLE derivations (1, 3, and 4) in Appendix~\ref{sec:appendixA}, the intuition for the derivation of multiplicative updates for $(D(\cdot \| \cdot),\|\cdot\|_F)$-SSNMF and $(D(\cdot \| \cdot),D(\cdot \| \cdot))$-SSNMF (Algorithms~\ref{algo:ifssnmf} and \ref{algo:iissnmf}) in Appendix~\ref{sec:appendixB}, and additional experimental results on the 20 Newsgroups data set in Appendix~\ref{ap:20news}.

\section{MLE Derivations}\label{sec:appendixA}
We begin by demonstrating that the MLE, in the case that the uncertainty on the $\mat X$ and $\mat Y$ observations is Gaussian distributed, is a specific instance of $(\|\cdot\|_F,\|\cdot\|_F)$-SSNMF of \cite{lee2009semi}. 
Our models for the distribution of the observed entries of $\mat X$ and $\mat Y$ will assume that the mean is given by an exact factorization, $\mathbb{E}[\mat X] = \mat A\mat S$ and $\mathbb{E}[\mat Y] = \mat B\mat S$, and the uncertainty in each set of observations is governed by a Gaussian distribution.  
That is, we consider the hierarchical models for $\mat X$ and $\mat Y$ in which $$\mat X_{\gamma,\tau} = \sum_{i=1}^r x_{\gamma,i,\tau} \text{ and } x_{\gamma,i,\tau} \sim \mathcal{N}\pars{x_{\gamma,i,\tau} \middle| \mat A_{\gamma,i}\mat S_{i,\tau}, \sigma_1},$$
\vspace{-1.5mm}
$$\mat Y_{\eta,\tau} = \sum_{i=1}^r y_{\eta,i,\tau} \text{ and } y_{\eta,i,\tau} \sim \mathcal{N}\pars{y_{\eta,i,\tau} \middle| \mat B_{\eta,i}\mat S_{i,\tau}, \sigma_2}.$$ 
Here and throughout, $\gamma$ and $\eta$ are row indices of $\mat X$ and $\mat Y$ respectively, $\tau$ is a column index of $\mat X$ and $\mat Y$, and $i$ indexes the random variable summands which form $\mat X_{\gamma,\tau}$ and $\mat Y_{\eta,\tau}$. Note then that $$\mat X_{\gamma,\tau} \sim \mathcal{N}\pars{\mat X_{\gamma,\tau} \middle| \sum_{i=1}^r \mat A_{\gamma,i}\mat S_{i,\tau}, r\sigma_1}, \mbox{  and}$$ $$\mat Y_{\eta,\tau} \sim \mathcal{N}\pars{\mat Y_{\eta,\tau} \middle| \sum_{i=1}^r \mat B_{\eta,i}\mat S_{i,\tau}, r\sigma_2}$$ due to the summable property of Gaussian random variables.  We note that this assumes different Gaussian models of uncertainty on the two collections of rows of the NMF \eqref{eq:stackedNMF}.  

Assuming that the set of $\mat X_{\gamma,\tau}$ and $\mat Y_{\eta,\tau}$ are statistically independent conditional on $\mat A$, $\mat B$, and $\mat S$, we have that the likelihood $p(\mat X,\mat Y|\mat A,\mat B,\mat S)$ is 

{\footnotesize%
\begin{equation}
    \prod_{\gamma,\tau} \mathcal{N}\pars{\mat X_{\gamma,\tau} \middle| \sum_{i=1}^r \mat A_{\gamma,i}\mat S_{i,\tau}, r\sigma_1} \prod_{\eta,\tau} \mathcal{N}\pars{\mat Y_{\eta,\tau} \middle| \sum_{i=1}^r \mat B_{\eta,i}\mat S_{i,\tau}, r\sigma_2}. \label{eq:frofrolikelihood}
\end{equation}}\normalsize

\noindent We apply the monotonic natural logarithmic function to the likelihood, and ignore terms that do not vary with the factor matrices.  This transforms the likelihood function into a $(\|\cdot\|_F,\|\cdot\|_F)$-SSNMF objective, while preserving the maximizer.
That is, the log likelihood (excluding additive terms which are constant with respect to $\mat A$, $\mat B$, and $\mat S$) is 

{\footnotesize%
\begin{align*}
\ln p\pars{\mat X, \mat Y \middle| \mat A, \mat B, \mat S} &=^+ -\frac{1}{2r\sigma_1} \sum_{\gamma,\tau}\pars{\mat X_{\gamma,\tau} - \sum_{i=1}^r \mat A_{\gamma,i}\mat S_{i,\tau}}^2 
\\&\hspace{1.2cm}- \frac{\lambda}{2r\sigma_2} \sum_{\eta,\tau} \pars{\mat Y_{\eta,\tau} - \sum_{i=1}^r \mat B_{\eta,i}\mat S_{i,\tau}}^2
\\&=^+ -\frac{1}{2r\sigma_1} \bracs{\|\mat X - \mat A\mat S\|_F^2 + \frac{\sigma_1}{\sigma_2} \|\mat Y - \mat B\mat S\|_F^2}.
\end{align*}}%
Thus, the maximum likelihood estimators for $\mat A$, $\mat B$, and $\mat S$ are given by 
$$\argmin_{\mat A,\mat B,\mat S \ge 0} \|\mat X - \mat A\mat S\|_F^2 + \frac{\sigma_1}{\sigma_2} \|\mat Y - \mat B\mat S\|_F^2.$$  We see that the MLE in the case of Gaussian uncertainty on both sets of observations, $\mat X$ and $\mat Y$, is a specific instance of $(\|\cdot\|_F,\|\cdot\|_F)$-SSNMF objective where the regularization parameter $\lambda$, which defines the relative weighting of the supervision term, is given as a ratio of the variances of the distributions.

Next, 
we demonstrate that the MLE, in the case that the uncertainty on $\mat X$ is Poisson distributed and on $\mat Y$ is Gaussian distributed, is a specific instane of the $(D(\cdot\|\cdot),\|\cdot\|_F)$-SSNMF model. This MLE derivation follows from that of 2 by swapping the roles of $\mat X$ and $\mat Y$, and rescaling the resulting log likelihood; however, we include a sketch of the derivation to be thorough.  

Again, our models for observed $\mat X$ and $\mat Y$ assume that the mean is given by an exact factorization, $\mathbb{E}[\mat X] = \mat A\mat S$ and $\mathbb{E}[\mat Y] = \mat B\mat S$, with the uncertainty in $\mat X$ governed by a Poisson distribution and the uncertainty in $\mat Y$ governed by a Gaussian distribution.  That is, we consider the hierarchical models for $\mat X$ and $\mat Y$ in which $$\mat X_{\gamma,\tau} = \sum_{i=1}^r x_{\gamma,i,\tau} \text{ and } x_{\gamma,i,\tau} \sim \mathcal{PO}\pars{x_{\gamma,i,\tau} \middle| \mat A_{\gamma,i}\mat S_{i,\tau}},$$ 
$$\mat Y_{\eta,\tau} = \sum_{i=1}^r y_{\eta,i,\tau} \text{ and } y_{\eta,i,\tau} \sim \mathcal{N}\pars{y_{\eta,i,\tau} \middle| \mat B_{\eta,i}\mat S_{i,\tau},\sigma_2}.$$ Note then that $$\mat X_{\gamma,\tau} \sim \mathcal{PO}\pars{\mat X_{\gamma,\tau} \middle| \sum_{i=1}^r \mat A_{\gamma,i}\mat S_{i,\tau}}, \mbox{  and}$$ $$\mat Y_{\eta,\tau} \sim \mathcal{N}\pars{\mat Y_{\eta,\tau} \middle| \sum_{i=1}^r \mat B_{\eta,i}\mat S_{i,\tau},r\sigma_2}$$ due to the summable property of Gaussian and Poisson random variables.  We note this assumes a Poisson and Gaussian model of uncertainty on the two collections of rows of the NMF \eqref{eq:stackedNMF}. 

Then proceeding as in \eqref{eq:frofrolikelihood} and \eqref{eq:frodivlikelihood} and assuming that the set of $\mat X_{\gamma,\tau}$ and $\mat Y_{\eta,\tau}$ are statistically independent conditional on $\mat A$, $\mat B$, and $\mat S$, we have that the 
log likelihood (excluding additive terms which are constant with respect to $\mat A$, $\mat B$, and $\mat S$) is 

{\footnotesize%
\begin{align*}
\ln p\pars{\mat X, \mat Y \middle| \mat A, \mat B, \mat S} 
&=^+ -\bracs{D(\mat X\|\mat A\mat S) + \frac{1}{2r\sigma_2} \|\mat Y-\mat B\mat S\|_F^2}.
\end{align*}}%
Thus, the maximum likelihood estimators for $\mat A$, $\mat B$, and $\mat S$ are given by 
$$\argmin_{\mat A,\mat B,\mat S \ge 0} D(\mat X\|\mat A\mat S) + \frac{1}{2r\sigma_2} \|\mat Y - \mat B\mat S\|_F^2.$$  We see that the MLE in the case of Poisson uncertainty on the observations in $\mat X$ and Gaussian uncertainty on the observations in $\mat Y$ is a specific instance of the $(D(\cdot\|\cdot),\|\cdot\|_F)$-SSNMF objective where the regularization parameter $\lambda$ is the inverse of a multiple of the variance of the Gaussian distribution.

Finally, we demonstrate that the MLE, in the case that the uncertainty on $\mat X$ and $\mat Y$ are Poisson distributed, is a specific instance of the $(D(\cdot\|\cdot),D(\cdot\|\cdot))$-SSNMF model.  This result follows from~\cite{cemgil2008bayesian,favaro20073,virtanen2008bayesian}; we sketch the derivation to be thorough.  

Again, we assume that the distributions of the observed  $X$ and $Y$ have means given by an exact factorization, $\mathbb{E}[\mat X] = \mat A\mat S$ and $\mathbb{E}[\mat Y] = \mat B\mat S$, with the uncertainty in both governed by a Poisson distribution.  That is, we consider the hierarchical models for $\mat X$ and $\mat Y$ in which $$\mat X_{\gamma,\tau} = \sum_{i=1}^r x_{\gamma,i,\tau} \text{ and } x_{\gamma,i,\tau} \sim \mathcal{PO}\pars{x_{\gamma,i,\tau} \middle| \mat A_{\gamma,i}\mat S_{i,\tau}},$$ 
$$\mat Y_{\eta,\tau} = \sum_{i=1}^r y_{\eta,i,\tau} \text{ and } y_{\eta,i,\tau} \sim \mathcal{PO}\pars{y_{\eta,i,\tau} \middle| \mat B_{\eta,i}\mat S_{i,\tau}}.$$ Note then that $$\mat X_{\gamma,\tau} \sim \mathcal{PO}\pars{\mat X_{\gamma,\tau} \middle| \sum_{i=1}^r \mat A_{\gamma,i}\mat S_{i,\tau}}, \mbox{  and}$$ $$\mat Y_{\eta,\tau} \sim \mathcal{PO}\pars{\mat Y_{\eta,\tau} \middle| \sum_{i=1}^r \mat B_{\eta,i}\mat S_{i,\tau}}$$ due to the summable property of Poisson random variables.  We note that assumes different Poisson models of uncertainty on the two collections of rows of the NMF \eqref{eq:stackedNMF}. 

Then proceeding as in \eqref{eq:frofrolikelihood} and \eqref{eq:frodivlikelihood} and assuming 
that the set of $\mat X_{\gamma,\tau}$ and $\mat Y_{\eta,\tau}$ are statistically independent conditional on $\mat A$, $\mat B$, and $\mat S$, we have that the 
log likelihood (excluding additive terms which are constant with respect to $\mat A$, $\mat B$, and $\mat S$) is 
\begin{align*}
\ln p\pars{\mat X, \mat Y \middle| \mat A, \mat B, \mat S} &=^+ 
-\bracs{D(\mat X\|\mat A\mat S) + D(\mat Y\| \mat B\mat S)}.
\end{align*}
Thus, the maximum likelihood estimators for $\mat A$, $\mat B$, and $\mat S$ are given by 
$$\argmin_{\mat A,\mat B,\mat S \ge 0} D(\mat X\|\mat A\mat S) + D(\mat Y\|\mat B\mat S).$$  We see that the MLE in the case of Poisson uncertainty on the observations in $\mat X$ and $\mat Y$ is a specific instance of the $(D(\cdot\|\cdot),D(\cdot\|\cdot))$-SSNMF objective where the regularization parameter is $\lambda = 1$.

\section{Multiplicative Updates Derivations}\label{sec:appendixB}
Here we provide intuition for the derivation of multiplicative updates for $(D(\cdot \| \cdot),\|\cdot\|_F)$-SSNMF and $(D(\cdot \| \cdot),D(\cdot \| \cdot))$-SSNMF (Algorithms~\ref{algo:ifssnmf} and \ref{algo:iissnmf}).

First, note that the multiplicative updates for $(D(\cdot \| \cdot),\|\cdot\|_F)$-SSNMF (Algorithm~\ref{algo:ifssnmf}) follow from those for $(\|\cdot\|_F,D(\cdot \| \cdot))$-SSNMF (Algorithm~\ref{algo:fissnmf}) by swapping the roles of $\mat X$ and $\mat Y$, and $\mat A$ and $\mat B$.

Next, the multiplicative updates for $(D(\cdot \| \cdot), D(\cdot, \| \cdot)$-SSNMF (Algorithm~\ref{algo:iissnmf}) 
are derived as follows.  
The gradients of $F_4(\mat A, \mat B, \mat S; \mat X, \mat Y)$ with respect to $\mat A$, $\mat B$, and $\mat S$ are, respectively
\begin{align*}
 \nabla_{\mat A} F_4 & = \mat W\mat S^\top - \left [ \frac{\mat W \odot \mat X}{\mat W \odot \mat A\mat S} \odot \mat W \right ]\mat S^\top, \\ \nabla_{\mat B} F_4 & = \mat L\mat S^\top - \left [ \frac{\mat L \odot \mat Y}{\mat L \odot \mat B\mat S} \odot \mat L \right ]\mat S^\top, \mbox{ and}\\
  \nabla_{\mat S} F_4 & =  -\mat A^\top \left[\frac{(\mat W \odot \mat X)}{(\mat W \odot \mat A\mat S)} \odot \mat W\right] + \mat A^\top \mat W \\
  &- \lambda \mat B^\top \left[\frac{(\mat L \odot \mat Y)}{(\mat L \odot \mat B\mat S)} \odot \mat L\right] + \lambda \mat B^\top \mat L. 
\end{align*}

The multiplicative updates of Algorithm \ref{algo:iissnmf} are given by 
\begin{align*}
   \mat A & \leftarrow \mat A - \Gamma \odot \nabla_{\mat A} F_4 \text{ \;\;when\;\; } \Gamma = \frac{\mat A}{\mat W\mat S^\top}, \\
   \mat B & \leftarrow \mat B - \Gamma \odot \nabla_{\mat B} F_4 \text{ \;\;when\;\; } \Gamma = \frac{\mat B}{\mat L\mat S^\top}, \mbox{ and} \\
    \mat S & \leftarrow \mat S - \Gamma \odot \nabla_{\mat S} F_4 \text{ \;\;when\;\; } \Gamma = \frac{\mat S}{\mat A^\top \mat W + \lambda \mat B^\top \mat L}.
\end{align*}

\section{Additional 20 Newsgroups Results}
\label{ap:20news}
In this section, we include additional analysis and results for the 20 Newsgroups data set.
First, we summarize in Table~\ref{table:model_parameters} the hyperparameters used for the methods described in Section~\ref{subsec:20newsdata}.
We select the hyperparameters that result in the highest average classification accuracy of the validation set. 
For the SSNMF models, we search over $tol \in \{10^{-4},10^{-3},10^{-2}\}$, and $\lambda \in \{10,10^2,10^3\}$, and for the NMF model, we search over $tol \in \{10^{-5},10^{-4},10^{-3},10^{-2}\}$.
\begin{table}[htb]
\footnotesize
    \centering
{ \renewcommand{\arraystretch}{1.15}
\begin{tabular}{ l  l }
            \hline
            Model &\qquad hyperparameters \\
            \hline
            $(\|\cdot\|_F,\|\cdot\|_F)$ &\qquad $tol = 10^{-4}$, $\lambda =10^2 $ \\
            $(\|\cdot\|_F,D(\cdot\|\cdot))$ &\qquad $tol = 10^{-4}$, $\lambda =10 $\\
            $(D(\cdot\|\cdot),\|\cdot\|_F)$ &\qquad $tol = 10^{-3}$, $\lambda =10^2 $ \\
            $(D(\cdot\|\cdot),D(\cdot\|\cdot))$ &\qquad $tol = 10^{-3}$, $\lambda =10^3 $ \\
            $\|\cdot\|_F$-NMF &\qquad $tol = 10^{-4}$\\
\hline
\end{tabular}}
\caption{Hyperparameter selection for NMF and SSNMF models by selecting the hyperparameters that result with the highest average classification accuracy of the validation set (over 10 trials).}
\label{table:model_parameters}
\end{table}

\begin{figure}[ht]
    \centering
     \begin{subfigure}[b]{\columnwidth}
     \centering
    \includegraphics[width=\columnwidth]{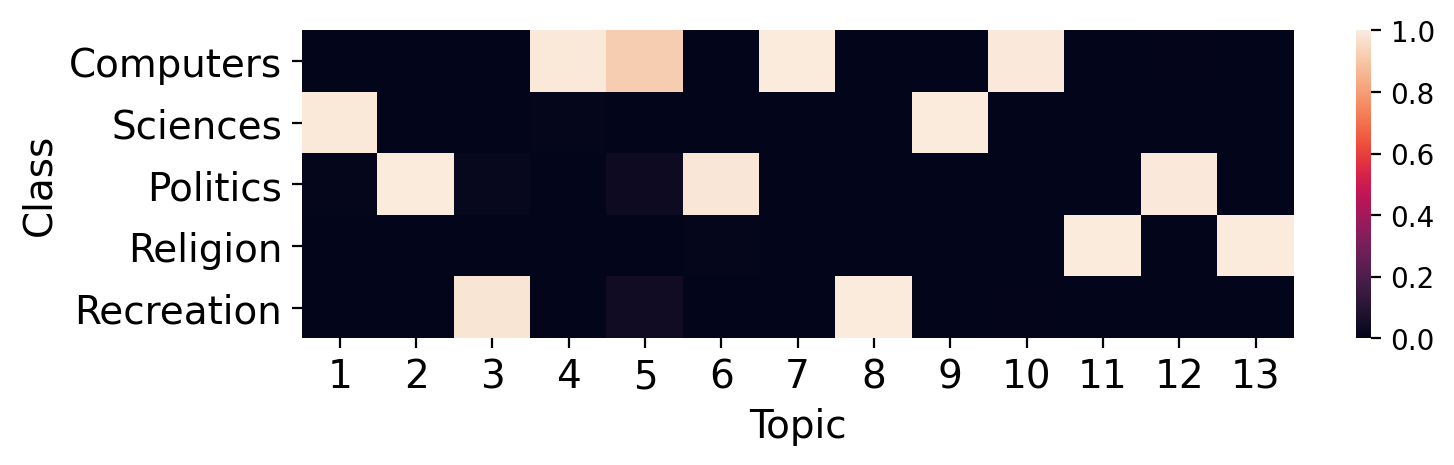}
     \caption{$(\|\cdot\|_F,\|\cdot\|_F)$-SSNMF decomposition corresponding to the median test classification accuracy equal to 79.44.}
     \label{fig:Model3_Normalized}
     \end{subfigure}
    \begin{subfigure}[b]{\columnwidth}
     \centering
    \includegraphics[width=\columnwidth]{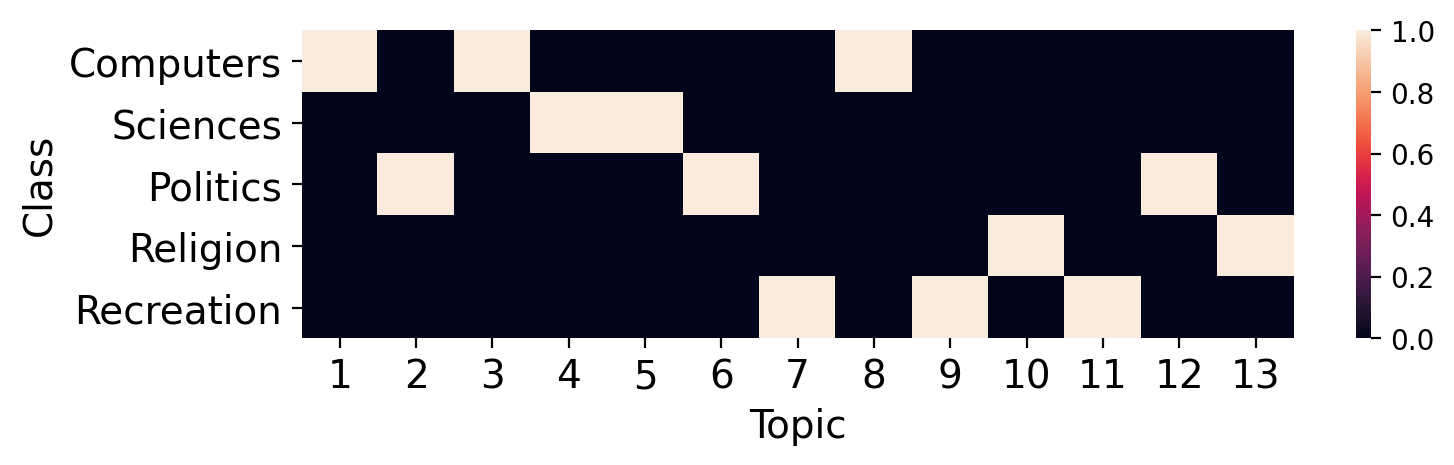}
     \caption{$(\|\cdot\|_F,D(\cdot\|\cdot))$-SSNMF decomposition corresponding to the median test classification accuracy equal to 79.56.}
     \label{fig:Model4_Normalized}
     \end{subfigure}
    \begin{subfigure}[b]{\columnwidth}
     \centering
    \includegraphics[width=\columnwidth]{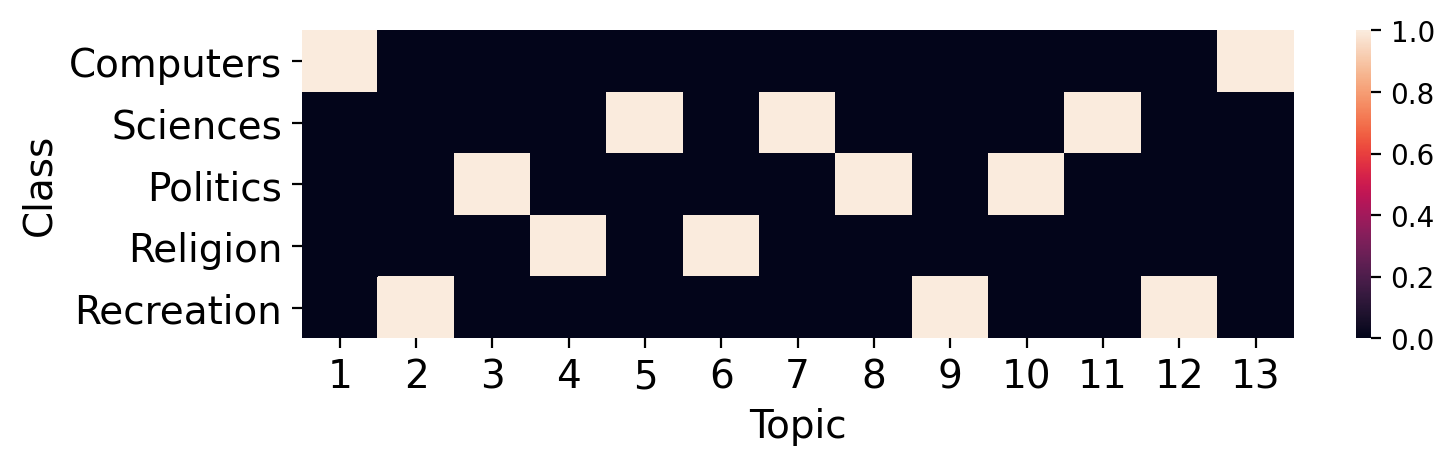}
     \caption{$(D(\cdot\|\cdot),D(\cdot\|\cdot))$-SSNMF decomposition corresponding to the median test classification accuracy equal to 81.39.}
     \label{fig:Model6_Normalized}
     \end{subfigure}
    \caption{The normalized $\mat B_{\text{train}}$ matrix for each of the SSNMF decomposition corresponding to the median test classification accuracy. Each column is normalized to represent the distribution of the topic over classes.}
    \label{fig:ssnmf_models_heatmap}
\end{figure}

\begin{figure}[ht]
    \centering
    \includegraphics[width=\columnwidth]{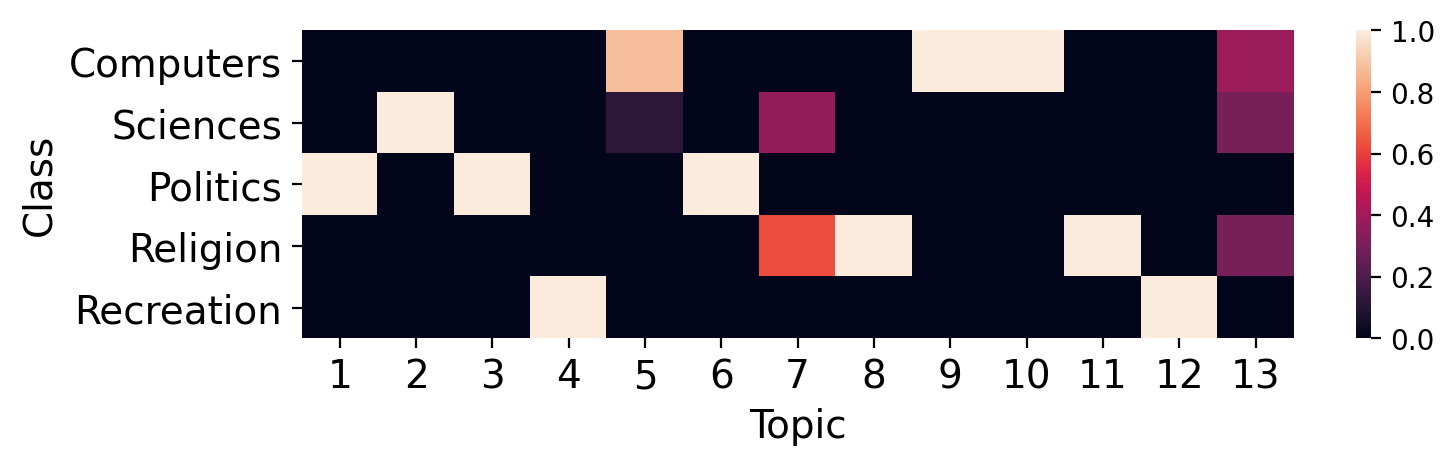}
    \caption{The normalized coefficients matrix of the SVM classifier (with NMF-$\|\cdot\|_F$) corresponding to the median test classification accuracy equal to 71.67. Here, all negative coefficients are thresholded to 0, and then each column is normalized to showcases the distribution of the topic over classes.}
    \label{fig:nmf_heatmap}
\end{figure}

As in Section~\ref{subsec:20newsdata}, we consider the ``typical" (achieving median accuracy within trials) decomposition for NMF, and remaining SSNMF models. 
We display in Figure~\ref{fig:ssnmf_models_heatmap} the $\mat B_{\text{train}}$ matrices for each of the median accuracy SSNMF decompositions, and in Figure~\ref{fig:nmf_heatmap} the coefficients matrix of the SVM classifier for the median accuracy NMF decomposition.
Further, we report in Tables~\ref{table:model3_keywords}, \ref{table:model4_keywords}, 
\ref{table:model6_keywords}, and \ref{table:nmf_keywords} the top 10 keywords representing each topic for each of the models.

For the $(\|\cdot\|_F,\|\cdot\|_F)$-SSNMF model, we (qualitatively) observe from Table~\ref{table:model3_keywords} that topics 4, 5, 7 and 10 are overlapping topics associated to the class Computers; see Figure~\ref{fig:Model3_Normalized}. 
Similarly, topics 1 and 9 that capture the subjects of crypt(ography), electronics, and space, have various overlapping keywords (``space", ``key",``chip"), and are associated with the class Sciences.
On the other hand, topics associated to class Politics and Religion are less overlapping.
Lastly, topics 3 and 8 are recreation topics (``game", ``team", ``car") relating to autos where in addition topic 3 (``hockey", ``player", ``nhl") is specific to hockey and topic 8 (``baseball") is specific to baseball.

For the $(\|\cdot\|_F,D(\cdot\|\cdot))$-SSNMF model, we (qualitatively) observe from Table~\ref{table:model4_keywords} that topic 2 (``armenian"), topic 6 (``jews",``gun", ``fire") and topic 12 (``guns") are related political topics (``people", ``government", ``israel", ``fbi"). 
The topics are associated to the class Politics; see Figure~\ref{fig:Model4_Normalized}.
Further, recreation topics include topic 7 (``player", ``hockey", ``baseball") relating to hockey and baseball, topic 9 (``car",``team") relating to autos, and a broad topic 11 (``game", ``good", ``great", ``play", ``better").
Lastly, topics 1, 3 and 8 are associated to class Computers. Topic 1 (``window",``graphics",``software",``windows"), and topic 3  (``mac", ``sun", ``graphics"), are specific in comparison to topic 8 (``ordinary",``yeah",``monitor") which is broad and not as informative.

For the $(D(\cdot\|\cdot),D(\cdot\|\cdot))$-SSNMF model, we observe from Figure~\ref{fig:Model6_Normalized} that topic 5 (``space", ``moon", ``time"), topic 7 (``key",``chip",``clipper"), and topic 11 (``data",``government",``buy") are associated with class Sciences.
Further, we (qualitatively) observe from Table~\ref{table:model6_keywords} that topic 4 (``church", ``believe",``bible"), and topic 6 (``atheists", ``paul") are both related to religion (``god", ``jesus") and are associated to class Religion; see Figure~\ref{fig:Model6_Normalized}.
Lastly, topic 12 (``car", ``hockey", ``players", ``baseball") is a recreation topic that captures autos, hockey, and baseball subjects, whereas topics 2 and 9 are broad and not as informative.

For the NMF-$\|\cdot\|_F$ model, we (qualitatively) observe from Table~\ref{table:nmf_keywords}, that topic 12 (``car", ``engine", ``oil") is related to autos, and topic 4 (``game", ``team", ``hockey",``baseball") captures other recreation games like hockey and baseball.
We observe in Figure~\ref{fig:nmf_heatmap} that topics 4 and 12 are associated to the class Recreation.
Further, topic 8 (``book",``true",``evidence") relates to atheism, and topic 11 (``god",``jesus",``christ",``faith") relates to religion and specifically Christianity.
Lastly, we observe in Figure~\ref{fig:nmf_heatmap} that topic 13 is a shared across 3 classes (Computers, Sciences, and Religion), and is not as informative as the other topics.
\begin{table*}[htb]
 \resizebox{\textwidth}{!}{
    \centering
\begin{tabular}{ |c|c|c|c|c|c|c|c|c|c|c|c|c|c|} 
\hline
\textbf{Topics} & Topic 1 & Topic 2 & Topic 3 & Topic 4 & Topic 5 & Topic 6 & Topic 7 & Topic 8 & Topic 9 & Topic 10 & Topic 11 & Topic 12 & Topic 13 \\
\hline
& would & people & game & use & x & would & x & game & would & please & god & israel & god \\
& like & israel & team & thanks & software & jews & thanks & car & key & x & church & people & one \\
& space & gun & car & x & c & fbi & get & team & could & would & jesus & would & would \\
& one & one & year & using & know & time & need & like & one & use & one & one & people \\
\textbf{Keywords} & chip & jews & hockey & window & r & israel & image & games & space & anyone & would & killed & jesus \\
& key & would & espn & know & thanks & like & window & baseball & use & like & people & armenians & believe \\
& use & armenian & would & graphics & widget & law & problem & one & chip & graphics & think & police & christ \\
& good & government & players & pc & system & arabs & windows & think & know & work & like & jewish & religion \\
& phone & said & nhl & program & please & government & mac & would & get & help & say & well & think \\
& edu & turkish & games & anyone & motif & right & version & get & like & apple & faith & israeli & bible \\
\hline
\textbf{Hard} & electronics & mideast & hockey & graphics & windows & guns & windows & baseball & crypt & graphics & christian & guns & christian \\
\hline
\textbf{Score} & 0.2060 & 0.6594 & 0.3411 & 0.2211 & 0.1047 & 0.1466 & 0.5698 & 0.7500 & 0.7641 & 0.1441 & 0.5226 & 0.1625 & 0.4574 \\
\hline
\textbf{Soft} & space & mideast & hockey & graphics & windows & guns & windows & baseball & crypt & graphics & christian & mideast & christian \\
\hline
\textbf{Score} & 0.3135 & 0.4560 & 0.4222 & 0.2389 & 0.1951 & 0.2278 & 0.3564 & 0.5933 & 0.6276 & 0.2128 & 0.4983 & 0.2480 & 0.4525 \\
\hline
\end{tabular}
}
\caption{Top keywords representing each topic of the $(\|\cdot\|_F,\|\cdot\|_F)$-SSNMF model referred to in Figure~\ref{fig:Model3_Normalized}.}
\label{table:model3_keywords}
\end{table*} 
\subsubsection*{Clustering Analysis}

In this section, we measure the performance of the NMF and SSNMF topic models in a clustering-motivated score.
In these experiments, we measure the similarity of ground-truth clusters, encoded by a given label matrix $\mat M$, to NMF/SSNMF computed clusters, encoded by the SSNMF/NMF representation matrix $\mat S$.
We denote by $\mat M$ the (column-wise) one-hot encoded label matrix which maps documents to the subgroups to which they belong\footnote{In the 20 Newsgroups data set, each document belongs to only one subgroup.}, $\mat M \in \{0,1\}^{13 \times 8980}$ (subgroups by documents).
We denote by $\mat S$ be the representation matrix computed by NMF/SSNMF, in which the $i$th row provides the association of each document with the $i$th topic.  

We employ two approaches to clustering or mixture assignment.  The first is \emph{hard clustering} in which the documents are assigned to a single cluster corresponding to computed topics.  In this approach, we apply a mask to the representation matrix, $\hat{\mat S} = \text{label}(\mat S)$, where $\text{label}(\cdot)$ assigns the largest entry of each column to 1 and all other entries to 0.  The second approach is \emph{soft clustering} in which the documents are assigned to a distribution of clusters corresponding to the topics.  In this approach, we normalize each of the columns of the representation matrix to produce $\hat{\mat S}$.  

Now, in either approach, we apply a metric $P$ which measures the association between the $\ell$th topic-documents association $\hat{\mat S}_\ell$ and the best ground truth subgroup-documents association, $\mat M_I$ (here $\mat M_I$ is the $I$th row of $\mat M$); that is, for topic $\ell$, we define $I$ as
\[I = \argmax _{\mat i} \frac{\|\hat{\mat S}_\ell \odot \mat M_i\|_1}{\|\mat M_i\|_1},\]
and define score $P$ for the $\ell$th topic as \[ P(\hat{\mat S}_\ell) = \frac{\|\hat{\mat S}_\ell \odot \mat M_I\|_1}{\|\mat M_I\|_1},\]
where $\|\cdot\|_1$ denotes the $\ell_1$-norm.
We note that this metric is similar to that of~\cite{xu2003document}; we use score $P$ instead as it allows us to measure clustering performance topic-wise.   
We also note that the learned topics of NMF and SSNMF methods need not be in one-to-one correspondence with the subgroups in Table~\ref{table:20news_data} as topics are also learnt for the classification task at hand.

Now, we present in Table~\ref{table:model_clustering} the average (averaged over topics) score $P$ for the representation matrices computed by each of the NMF/SSNMF models in both the hard-clustering and soft-clustering settings.  The scores $P$ for each topic (for both hard-clustering and soft-clustering) and the maximizing subgroup (indicated by $I$) are listed in the bottom four rows of the keyword table associated to each model; see the last four rows of 
Tables~\ref{table:model3_keywords}, \ref{table:model4_keywords}, \ref{table:model5_clusters}, \ref{table:model6_keywords}, and \ref{table:nmf_keywords}.

\begin{table}[tb]
\footnotesize
    \centering
{ \renewcommand{\arraystretch}{1.15}
\begin{tabular}{ l  c  c}
            \hline
            Model & Hard Clustering & Soft Clustering \\
            \hline
            $(\|\cdot\|_F,\|\cdot\|_F)$ &0.3895 &0.3647 \\
            $(\|\cdot\|_F,D(\cdot\|\cdot))$ &0.3857 &0.3703 \\
            $(D(\cdot\|\cdot),\|\cdot\|_F)$ &0.3874 &0.3553 \\
            $(D(\cdot\|\cdot),D(\cdot\|\cdot))$ &0.3874 &\textbf{0.3732} \\
            $\|\cdot\|_F$-NMF &\textbf{0.4348} &0.3080 \\
\hline
\end{tabular}}
\caption{Listed scores are average over 11 trials; in each trial, we average score $P$ across all topics.}
\label{table:model_clustering}
\end{table}

\begin{table*}[htb]
 \resizebox{\textwidth}{!}{
    \centering
\begin{tabular}{ |c|c|c|c|c|c|c|c|c|c|c|c|c|c|} 
\hline
\textbf{Topics} & Topic 1 & Topic 2 & Topic 3 & Topic 4 & Topic 5 & Topic 6 & Topic 7 & Topic 8 & Topic 9 & Topic 10 & Topic 11 & Topic 12 & Topic 13 \\
\hline
& x & israel & thanks & would & would & people & game & ordinary & car & god & game & people & god \\
& know & would & x & chip & one & would & player & yeah & team & jesus & good & gun & one \\
& would & people & anyone & use & key & jews & hockey & monitors & game & deleted & one & one & would \\
& use & government & get & clipper & could & israel & espn & ok & games & science & car & guns & people \\
\textbf{Keywords} & window & israeli & mac & space & space & gun & would & big & like & would & get & israel & jesus \\
& graphics & one & sun & government & like & fbi & baseball & know & year & moses & anyone & said & church \\
& please & fbi & graphics & key & know & one & new & shareware & get & post & great & armenian & think \\
& software & armenians & file & much & get & law & wings & way & would & passages & play & government & bible \\
& windows & armenian & one & get & use & fire & think & anyone & one & come & year & well & believe \\
& mac & also & please & people & chip & think & players & good & think & commandments & better & would & christian \\
\hline
\textbf{Hard} & windows & mideast & graphics & crypt & electronics & guns & hockey & graphics & autos & atheism & baseball & mideast & christian \\
\hline
\textbf{Score} & 0.6429 & 0.3548 & 0.3719 & 0.3272 & 0.6478 & 0.2784 & 0.2268 & 0.0067 & 0.7676 & 0.0463 & 0.0405 & 0.3865 & 0.9298 \\
\hline
\textbf{Soft} & windows & mideast & windows & crypt & electronics & guns & hockey & graphics & autos & christian & baseball & mideast & christian \\
\hline
\textbf{Score} & 0.5262 & 0.3315 & 0.3821 & 0.3973 & 0.5444 & 0.2988 & 0.2717 & 0.0659 & 0.5669 & 0.1296 & 0.1274 & 0.3363 & 0.8198 \\
\hline
\end{tabular}
}
\caption{Top keywords representing each topic of the $(\|\cdot\|_F,D(\cdot\|\cdot))$-SSNMF model referred to in Figure~\ref{fig:Model4_Normalized}.}
\label{table:model4_keywords}
\end{table*}
\begin{table*}[htb]
 \resizebox{\textwidth}{!}{
    \centering
\begin{tabular}{ |c|c|c|c|c|c|c|c|c|c|c|c|c|c|} 
\hline
\textbf{Topics} & Topic 1 & Topic 2 & Topic 3 & Topic 4 & Topic 5 & Topic 6 & Topic 7 & Topic 8 & Topic 9 & Topic 10 & Topic 11 & Topic 12 & Topic 13 \\
\hline
\textbf{Hard} & crypt & autos & christian & windows & guns & hockey & hockey & mideast & electronics & space & mideast & space & christian \\
\hline
\textbf{Score} & 0.5698 & 0.4167 & 0.7332 & 0.9867 & 0.5693 & 0.3046 & 0.3146 & 0.1976 & 0.3106 & 0.0692 & 0.2140 & 0.1115 & 0.2477 \\
\hline
\textbf{Soft} & crypt & autos & christian & windows & mideast & hockey & baseball & mideast & electronics & crypt & mideast & electronics & christian \\
\hline
\textbf{Score} & 0.3284 & 0.3260 & 0.5527 & 0.9181 & 0.4010 & 0.3006 & 0.3056 & 0.2689 & 0.2733 & 0.1577 & 0.2601 & 0.2217 & 0.3571 \\
\hline
\end{tabular}
}
\caption{Clustering results for topics of the $(D(\cdot\|\cdot),\|\cdot\|_F)$-SSNMF model referred to in Figure~\ref{fig:SSNMF_Model5}.}
\label{table:model5_clusters}
\end{table*}
\begin{table*}[htb]
 \resizebox{\textwidth}{!}{
    \centering
\begin{tabular}{ |c|c|c|c|c|c|c|c|c|c|c|c|c|c|} 
\hline
\textbf{Topics} & Topic 1 & Topic 2 & Topic 3 & Topic 4 & Topic 5 & Topic 6 & Topic 7 & Topic 8 & Topic 9 & Topic 10 & Topic 11 & Topic 12 & Topic 13 \\
\hline
& thanks & game & israel & god & would & god & key & would & game & people & one & car & x \\
& x & games & one & would & space & atheists & would & people & win & gun & like & team & anyone \\
& mac & would & arab & one & could & one & chip & government & etc & get & use & game & thanks \\
& know & think & people & jesus & u & people & clipper & gun & turbo & right & get & year & graphics \\
\textbf{Keywords} & would & get & government & church & use & would & could & said & know & jews & would & like & get \\
& problem & back & would & people & moon & think & space & israel & games & armenian & think & hockey & use \\
& please & hit & fire & believe & time & paul & know & one & get & well & data & players & window \\
& use & well & well & bible & old & jesus & one & jews & would & us & anyone & last & would \\
& one & one & like & think & one & know & using & armenians & cup & time & government & one & please \\
& se & like & israeli & christian & may & also & like & batf & find & armenians & buy & baseball & know \\
\hline
\textbf{Hard} & graphics & hockey & mideast & christian & crypt & atheism & crypt & guns & autos & mideast & electronics & baseball & windows \\
\hline
\textbf{Score} & 0.2982 & 0.0993 & 0.4487 & 0.8947 & 0.1246 & 0.0876 & 0.5897 & 0.1875 & 0.0545 & 0.3526 & 0.3256 & 0.8521 & 0.7475 \\
\hline
\textbf{Soft} & graphics & hockey & mideast & christian & crypt & christian & crypt & mideast & hockey & mideast & space & baseball & windows \\
\hline
\textbf{Score} & 0.3682 & 0.1746 & 0.3643 & 0.7596 & 0.2175 & 0.1989 & 0.4307 & 0.2602 & 0.1306 & 0.3252 & 0.3199 & 0.6653 & 0.6240 \\
\hline
\end{tabular}
}
\caption{Top keywords representing each topic of the $(D(\cdot\|\cdot),D(\cdot\|\cdot))$-SSNMF model referred to in Figure~\ref{fig:Model6_Normalized}.}
\label{table:model6_keywords}
\end{table*}
\begin{table*}[htb]
 \resizebox{\textwidth}{!}{
    \centering
\begin{tabular}{ |c|c|c|c|c|c|c|c|c|c|c|c|c|c|} 
\hline
\textbf{Topics} & Topic 1 & Topic 2 & Topic 3 & Topic 4 & Topic 5 & Topic 6 & Topic 7 & Topic 8 & Topic 9 & Topic 10 & Topic 11 & Topic 12 & Topic 13 \\
\hline
& people & key & armenian & game & thanks & israel & would & one & x & get & god & car & know \\
& gun & chip & armenians & team & please & jews & like & two & r & mac & jesus & like & anyone \\
& right & clipper & turkish & games & mail & israeli & think & book & window & use & christ & cars & need \\
& government & encryption & genocide & year & advance & arab & could & another & server & space & faith & engine & something \\
\textbf{Keywords} & fbi & keys & armenia & hockey & edu & jewish & make & true & windows & system & believe & good & like \\
& guns & algorithm & turks & baseball & e & arabs & say & point & motif & software & sin & new & us \\
& law & escrow & soviet & last & anyone & palestinian & church & evidence & display & drive & bible & v & anybody \\
& us & phone & turkey & players & list & palestinians & might & may & running & apple & us & price & heard \\
& think & government & muslim & season & send & peace & someone & thing & application & mhz & christians & oil & sure \\
& batf & number & russian & espn & address & israelis & something & word & sun & monitor & lord & dealer & program \\
\hline
\textbf{Hard} & guns & crypt & mideast & hockey & graphics & mideast & space & atheism & windows & mac & christian & autos & electronics \\
\hline
\textbf{Score} & 0.7080 & 0.5266 & 0.2576 & 0.7947 & 0.2647 & 0.3854 & 0.1655 & 0.2115 & 0.5781 & 0.6499 & 0.5436 & 0.6138 & 0.0864 \\
\hline
\textbf{Soft} & guns & crypt & mideast & hockey & graphics & mideast & christian & atheism & windows & mac & christian & autos & graphics \\
\hline
\textbf{Score} & 0.4408 & 0.3594 & 0.2025 & 0.5635 & 0.2149 & 0.2707 & 0.1426 & 0.1722 & 0.4083 & 0.4519 & 0.3427 & 0.4238 & 0.0929 \\
\hline
\end{tabular}
}
\caption{Top keywords representing each topic of the NMF-$\|\cdot\|_F$ + SVM model referred to in Figure~\ref{fig:nmf_heatmap}.}
\label{table:nmf_keywords}
\end{table*}

\end{document}